\documentclass[10pt,twocolumn,letterpaper]{article}

\usepackage{wacv}
\usepackage{times}
\usepackage{epsfig}
\usepackage{graphicx}
\usepackage{amsmath}
\usepackage{amssymb}
\usepackage{booktabs}
\usepackage{comment}
\def\wacvPaperID{95} % Enter the WACV Paper ID here

\wacvapplicationstrack % Uncomment this line if you are submitting to the Applications Track.

\wacvfinalcopy % *** Uncomment this line for the final submission

\ifwacvfinal
\usepackage[breaklinks=true,bookmarks=false]{hyperref}
\else
\usepackage[pagebackref=true,breaklinks=true,colorlinks,bookmarks=false]{hyperref}
\fi

\begin{document}

%%%%%%%%% TITLE
\title{The Weaknesses of Adversarial Camouflage in Overhead Imagery}

\author{Adam Van Etten\\
IQT Labs\\
{\tt\small avanetten@iqt.org}
%\and
}

\maketitle
\thispagestyle{empty}

%%%%%%%%% ABSTRACT
\begin{abstract}

Machine learning is increasingly critical for analysis of the ever-growing corpora of overhead imagery.  Advanced computer vision object detection techniques have demonstrated great success in identifying objects of interest such as ships, automobiles, and aircraft from satellite and drone imagery.  Yet relying on computer vision opens up significant vulnerabilities, namely, the susceptibility of object detection algorithms to adversarial attacks.  In this paper we explore the efficacy and drawbacks of adversarial camouflage in an overhead imagery context.  While a number of recent papers have demonstrated the ability to reliably fool deep learning classifiers and object detectors with adversarial patches, most of this work has been performed on relatively uniform datasets and only a single class of objects.  In this work we utilize the VisDrone dataset, which has a large range of perspectives and object sizes.  We explore four different object classes: bus, car, truck, van.  We build a library of 24 adversarial patches to disguise these objects, and introduce a patch translucency variable to our patches.  The translucency (or alpha value) of the patches is highly correlated to their efficacy. Further, we show that while adversarial patches may fool object detectors, the presence of such patches is often easily uncovered, with patches on average 24\% more detectable than the objects the patches were meant to hide.  This raises the question of whether such patches truly constitute camouflage. Source code is available at 
%\url{link_removed_for_blind_review}.
\url{https://github.com/IQTLabs/camolo}.

\end{abstract}

\section{Introduction} \label{sec:intro}

Computer vision algorithms are known to be susceptible to perturbations: A 2017 study summarized its findings like this: ``Given a state-of-the-art deep neural network classifier, we show the existence of a universal (image-agnostic) and very small perturbation vector that causes natural images to be misclassified with high probability''  \cite{univ_pert}. Most recent research has focused on image classification and shown impressive results (\eg \cite{univ_pert}, \cite{nat_style_camo}, \cite{robust_attacks}, \cite{adv_laser}) in inducing mis-classifications.  A smaller, though significant, body of work has been performed on segmentation (\eg \cite{adv_2017}, \cite{univ_pert}).  

There is also an increasing body of work around fooling advanced object detection systems.  Some works (\cite{attn_camo}) have shown an impressive ability to fool the YOLO \cite{yolo} family of detectors with simulated autonomous vehicle data.  
Other works have concluded that at least for video object detection of vehicles, adversarial examples are not a significant concern due to varying distances and angles \cite{adv_veh}.  

Of greatest relevance to this work are two recent papers that seek to fool YOLO.  First up, \cite{adv-yolo} demonstrated the ability to fool YOLO models trained to detect people in surveillance cameras - significantly lowering the accuracy of person detectors trained on the Inria \cite{inria} dataset with the imposition of an adversarial patch.  More recently,  \cite{aerial_camo}, adapted the codebase of \cite{adv-yolo} and demonstrated the ability to fool aircraft detectors with simulated patches trained on the DOTA dataset \cite{dota}.  While these simulated patches of \cite{aerial_camo} may be quite effective in fooling aircraft detectors, they have not been tested in the real world or even projected in a realistic matter onto the objects of interest.  Both papers used only single-class models and datasets with relatively little variability in object size and viewing perspective.
 
In Section \ref{sec:motivation} we detail what motivates why we seek to expand upon the above prior work.  Section \ref{sec:data} details the drone imagery dataset used in this study, while Section \ref{sec:algo} describes our algorithmic approach.  In Section \ref{sec:experiments} we cover the results of our experiments, and in Section \ref{sec:analysis} we analyze the meaning of these results.

\section{Motivation}\label{sec:motivation}

\begin{figure}%[t]
\begin{center}
   \includegraphics[width=0.9\linewidth]{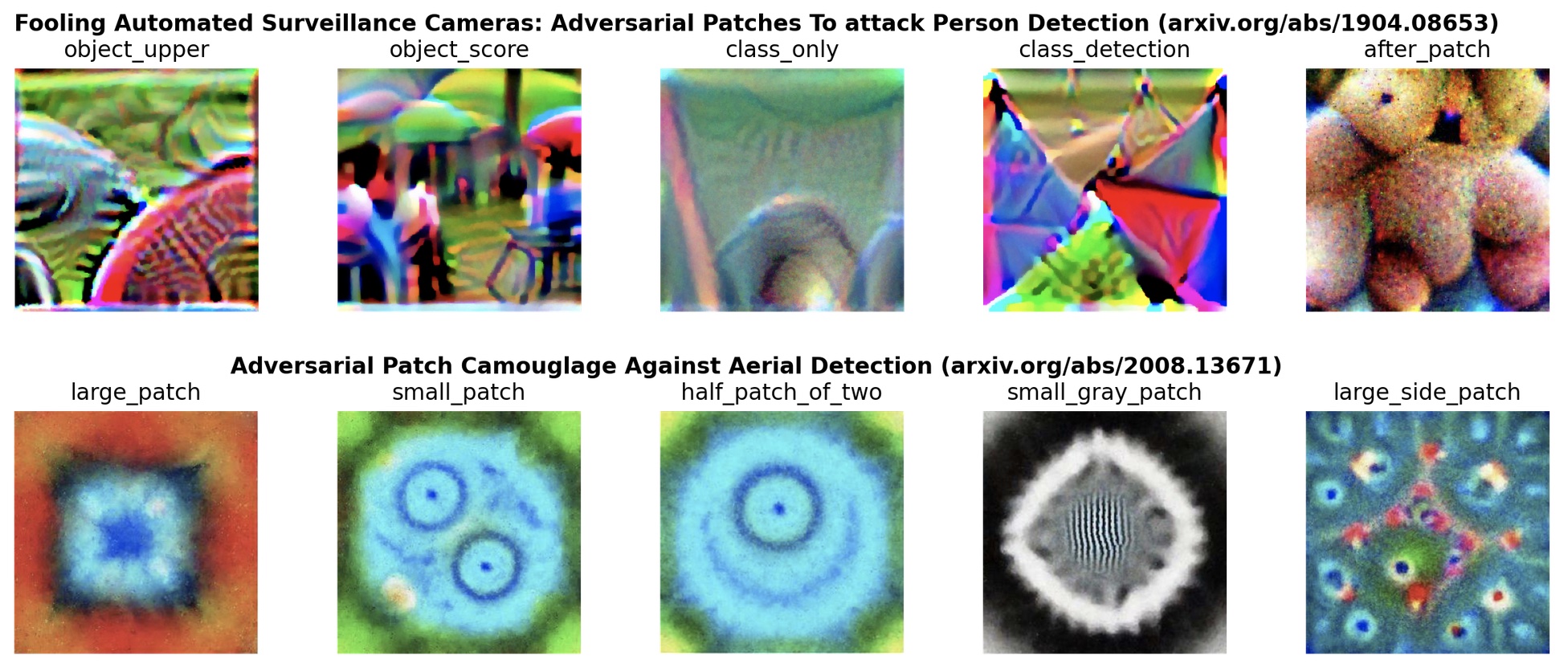}
\end{center}
  %\vspace{-5pt}
   \caption{Legacy adversarial patches. Top: \cite{adv-yolo}, bottom: \cite{aerial_camo}.}
\label{fig:legacy_patches}
\end{figure}

There is much to explore in the adversarial camouflage space, but we will start with a simple mitigation example.  In Section \ref{sec:intro} we summarized prior work on effectively obfuscating objects with carefully designed patches.  We ask a simple question: can the presence of adversarial patches be detected? To answer this question, we select a sampling of satellite image tiles (from the RarePlanes \cite{rareplanes} dataset) and overlay a 10 existing performant adversarial patches on these image tiles (see Figure \ref{fig:legacy_patches}). % for the patches utilized.

\begin{figure}%[t]
\begin{center}
   \includegraphics[width=0.9\linewidth]{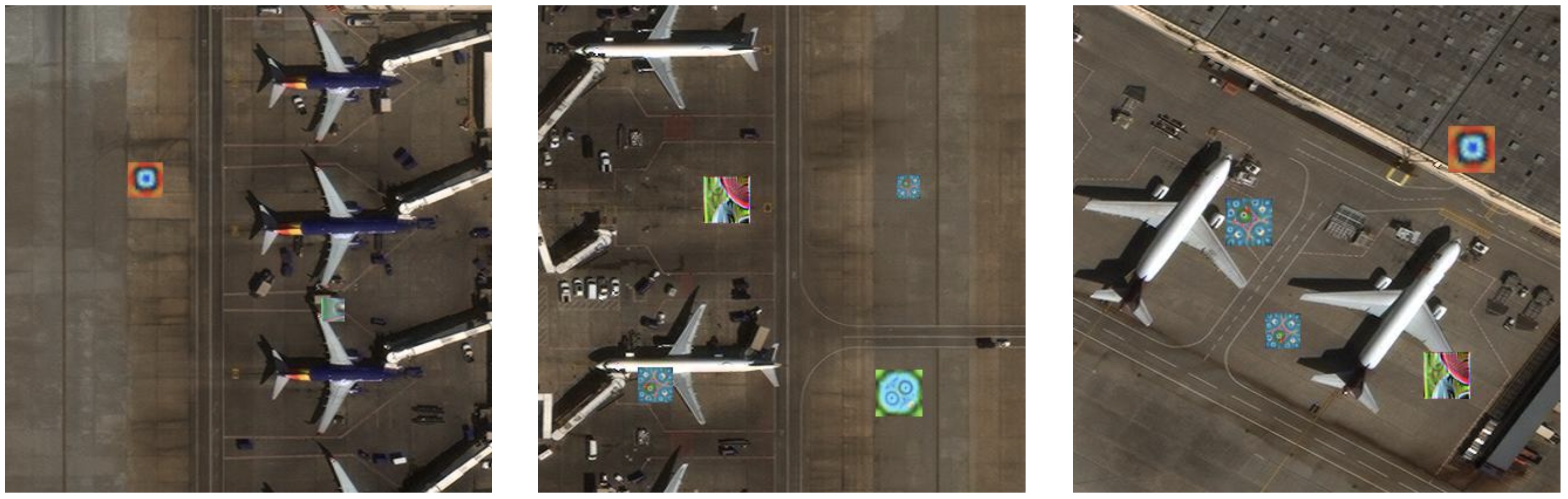}
\end{center}
  %\vspace{-5pt}
   \caption{Training data for our YOLTv4 patch detection model.}
\label{fig:patch_overlays}
\end{figure}

\begin{figure}%[t]
\begin{center}
   \includegraphics[width=0.9\linewidth]{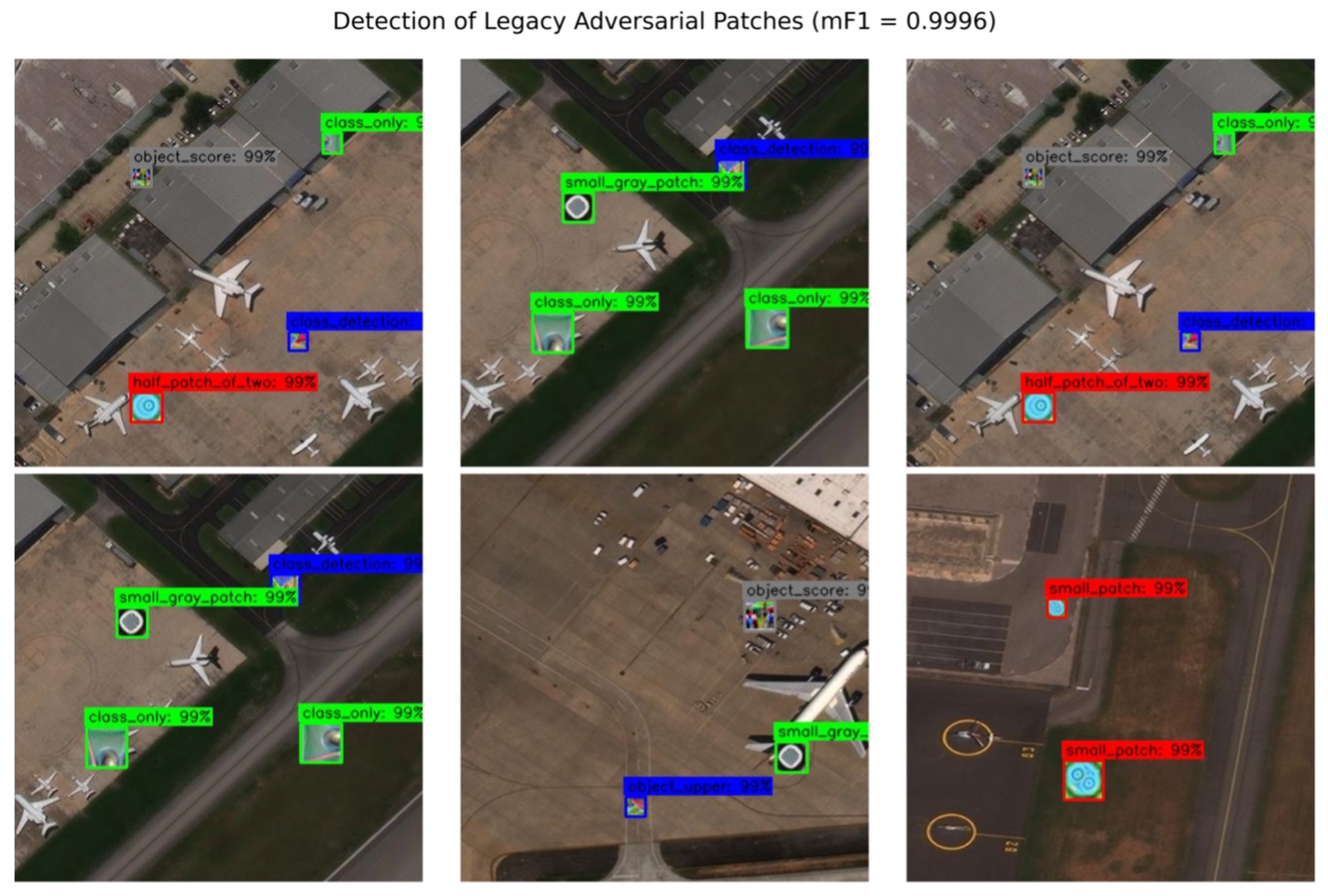}
\end{center}
  %\vspace{-5pt}
   \caption{Detection of the presence of adversarial patches with our YOLTv4 model.}
\label{fig:patch_det}
\end{figure}

We overlay the ten patches of Figure \ref{fig:legacy_patches} on 1667 image patches from the RarePlanes dataset, with 20\% of these images reserved for validation.  See Figure \ref{fig:patch_overlays} for a sample of the training imagery. We train a 10-class YOLTv4 model  \cite{yoltv4} (YOLTv4 is built atop YOLO \cite{yolo} and designed for overhead imagery analysis) for 20 epochs.  We evaluate our trained model on a separate test set with 1029 images, and score with the F1 metric that penalizes both false positives and false negatives. The model yields an astonishing aggregate F1 = 0.999 for detection of the adversarial patches, with all 10 patches easily identified.  We show a detection example in Figure \ref{fig:patch_det}. 

If one can easily identify the existence of adversarial patches, the real-world effectiveness of these patches for camouflaging objects is called into question.  One could simply run a patch detector in parallel with the original object detector.  This motivates our study in Section \ref{sec:algo} on whether ``stealthy'' patches can be designed that are both difficult to detect, and effective at obfuscating the presence of objects of interest.

\section{VisDrone Dataset}\label{sec:data}

\begin{figure}%[t]
\begin{center}
   \includegraphics[width=0.9\linewidth]{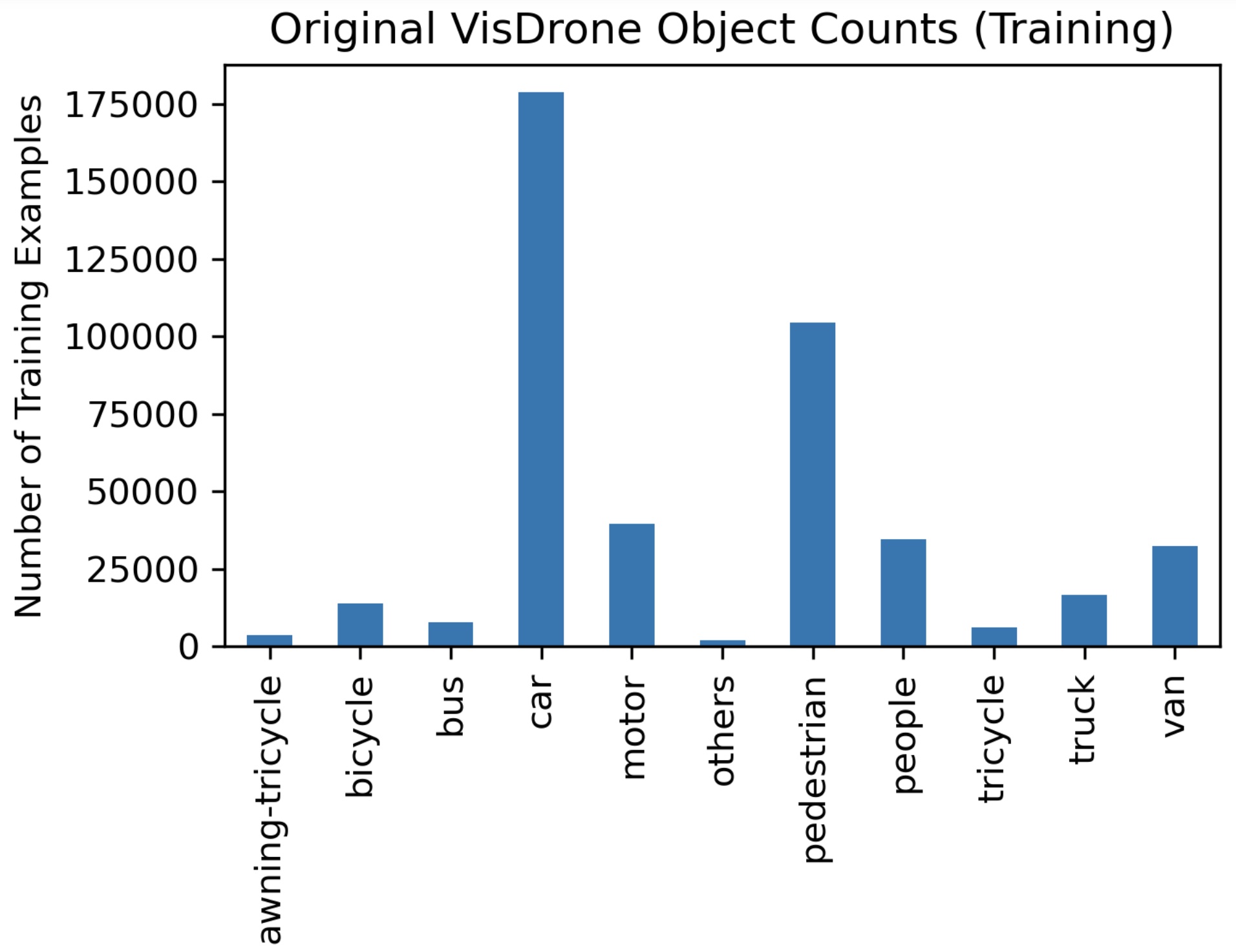}
\end{center}
  %\vspace{-5pt}
   \caption{Object class counts for the raw VisDrone training dataset.}
\label{fig:histo_orig}
\end{figure}

\begin{figure}
\begin{tabular}{cc}
  \includegraphics[width=0.48\linewidth]{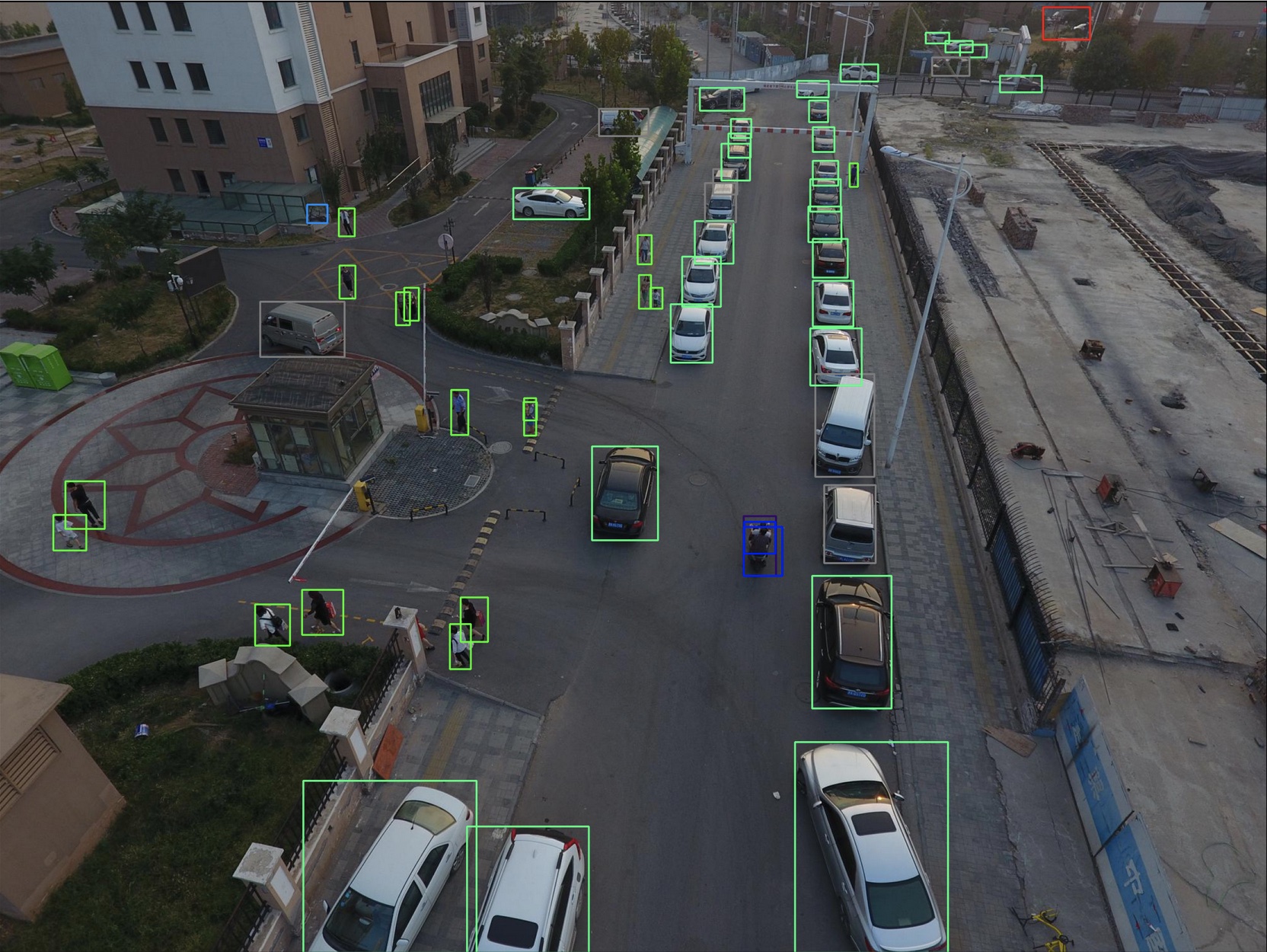} &   
  \includegraphics[width=0.48\linewidth]{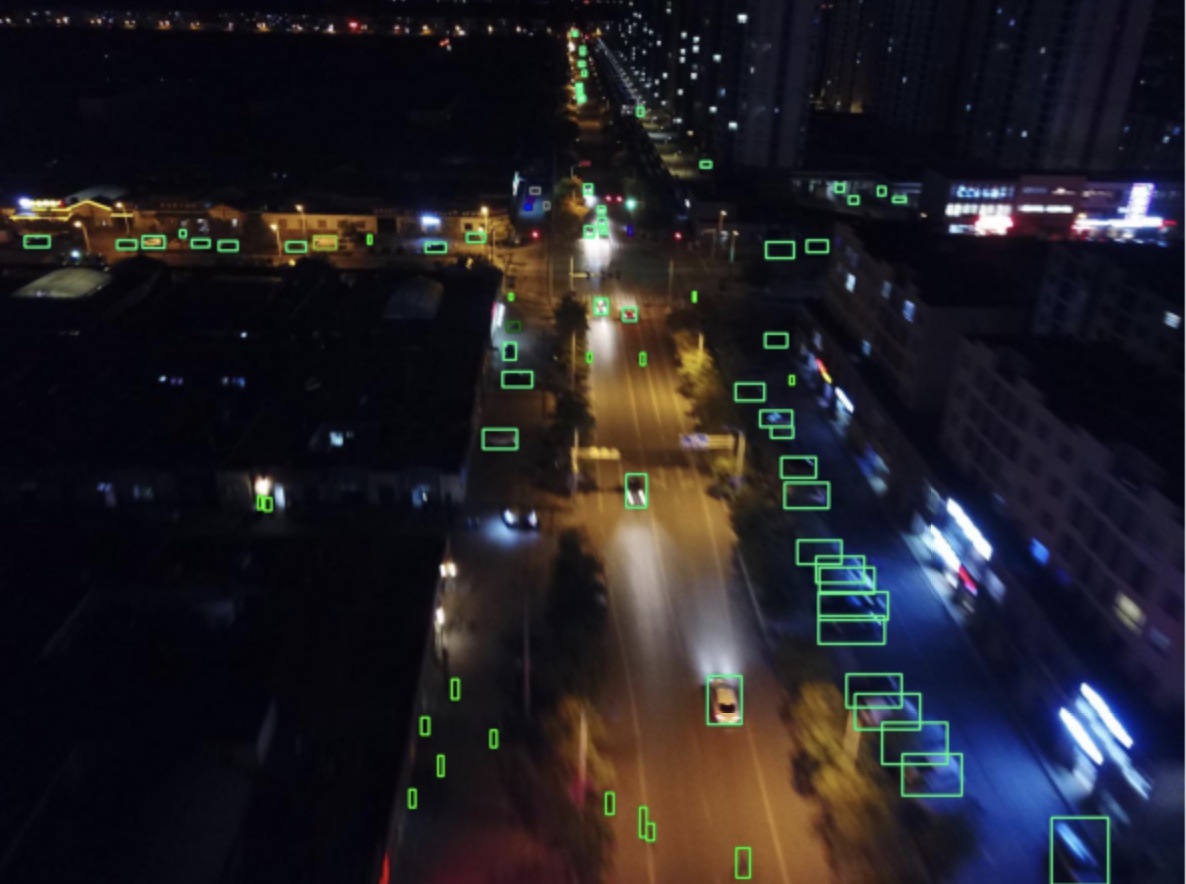} \\ 
  [-2pt] 
%(a) Raw Image & (b) UDM overlaid \\  [4pt] 
 \includegraphics[width=0.48\linewidth]{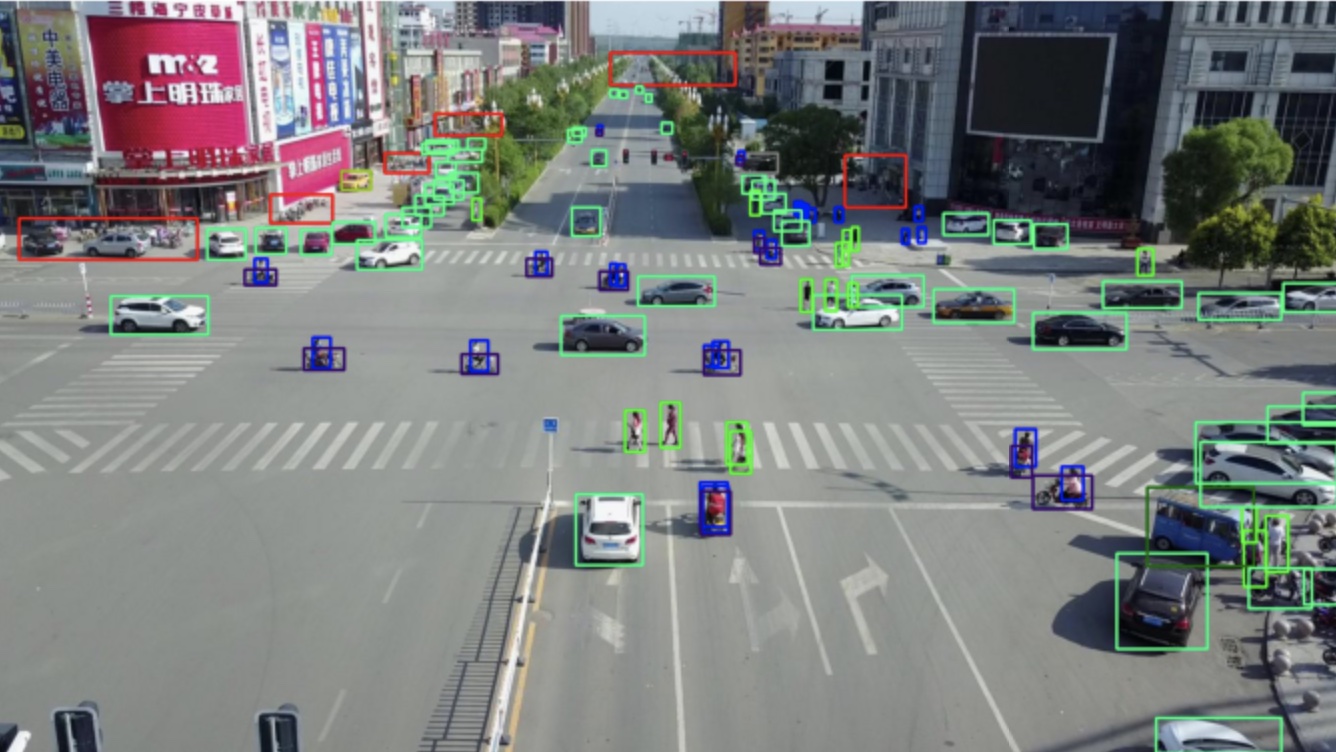} &  
 \includegraphics[width=0.48\linewidth]{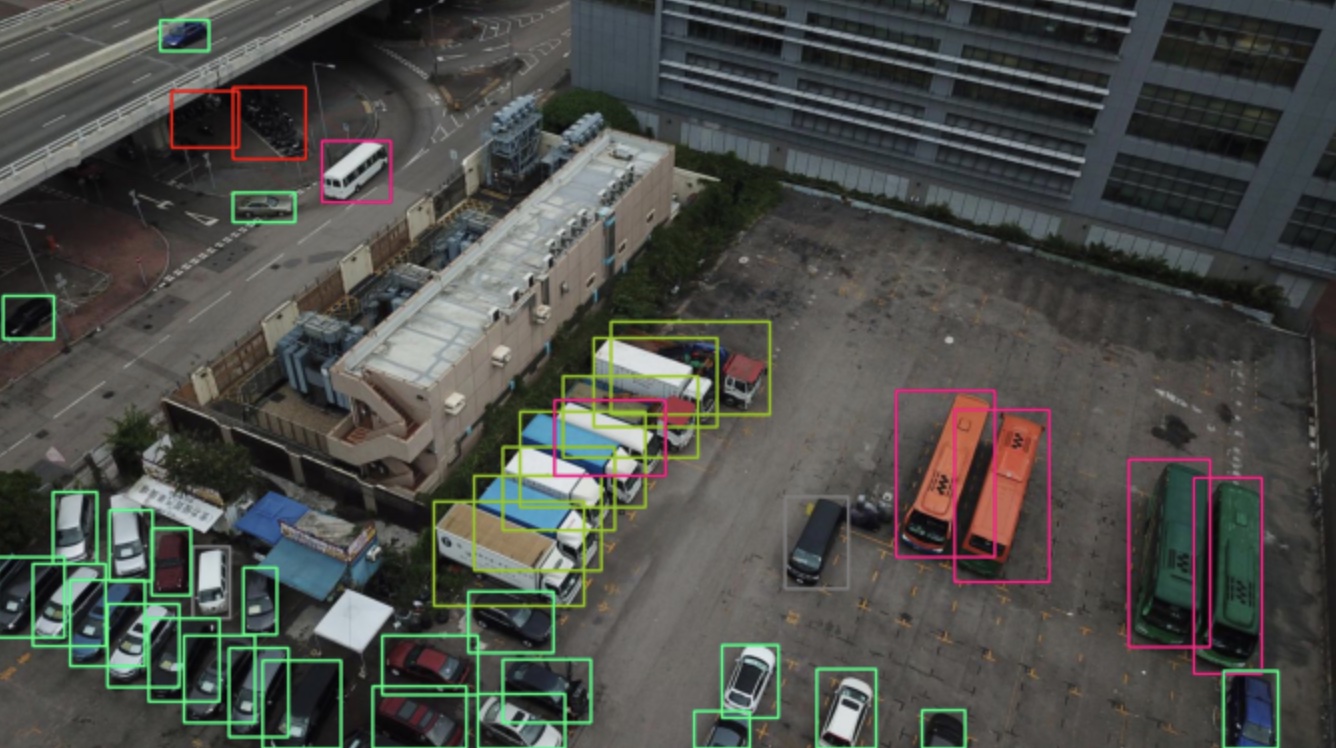} \\  
% [-2pt] 
%(c) Masked image + labels & (d) Zoomed labels \\  [-0pt] 
% \multicolumn{2}{c}{\includegraphics[width=0.99\linewidth]{it} }\\
% \multicolumn{2}{c}{(e) fifth}
\end{tabular}
\caption{Sample images and ground truth bounding box labels for the raw VisDrone training dataset.  Red boxes denote unlabeled regions.}
\label{fig:raw_boxes}
%\vspace{-1pt}
\end{figure}

For the remainder of this paper we use the VisDrone \cite{visdrone} dataset, specifically the object detection portion of VisDrone (VisDrone2019-DET).  This dataset includes 6471 images taken from drones in the training set, with 1610 images in the test set.  Bounding box annotations are provided for 11 object classes: pedestrian, person, car, van, bus, truck, motor, bicycle, awning-tricycle, tricycle and ``others.''  The dataset is highly imbalanced: ``car'' and ``pedestrian'' are by far the most common classes, see Figure \ref{fig:histo_orig}, and a high number of labels per image (median of 43 labels per image, max of 914 labels).  The altitude, viewing angle, and lighting conditions are highly variable, which complicates analysis of the imagery (see Figure \ref{fig:raw_boxes}).  The 353,550 bounding box labels in the training set tend to be relatively small (median extent of 34 pixels), though the size is highly variable (std of 44 pixels), which is a due to the variance in altitude and viewing angle of the drone platform.

\begin{figure}%[t]
\begin{center}
   \includegraphics[width=0.9\linewidth]{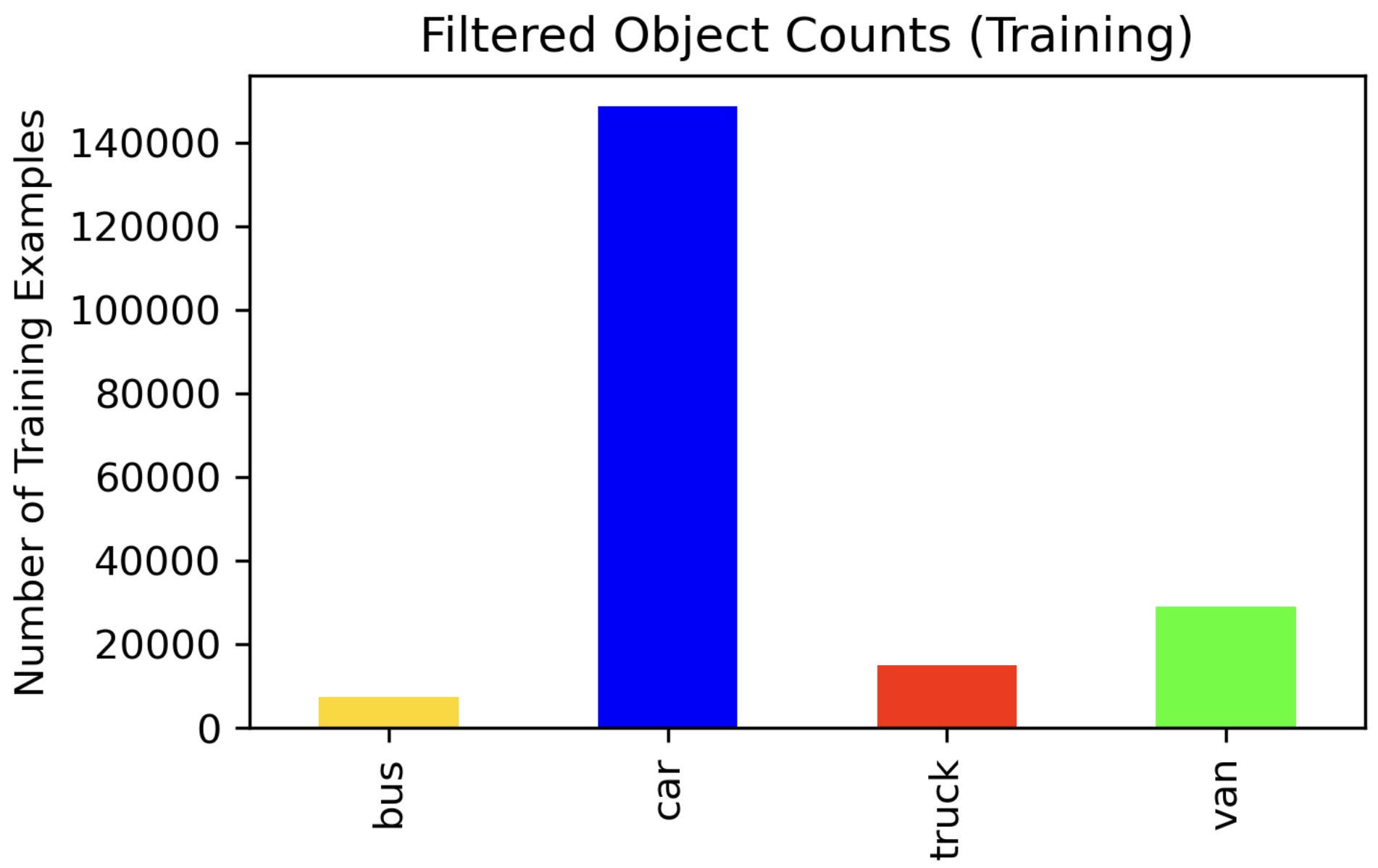}
\end{center}
  %\vspace{-5pt}
   \caption{Object class counts for our filtered 4-class VisDrone training dataset.}
\label{fig:histo_filt}
\end{figure}

\begin{figure}%[t]
\begin{center}
   \includegraphics[width=0.9\linewidth]{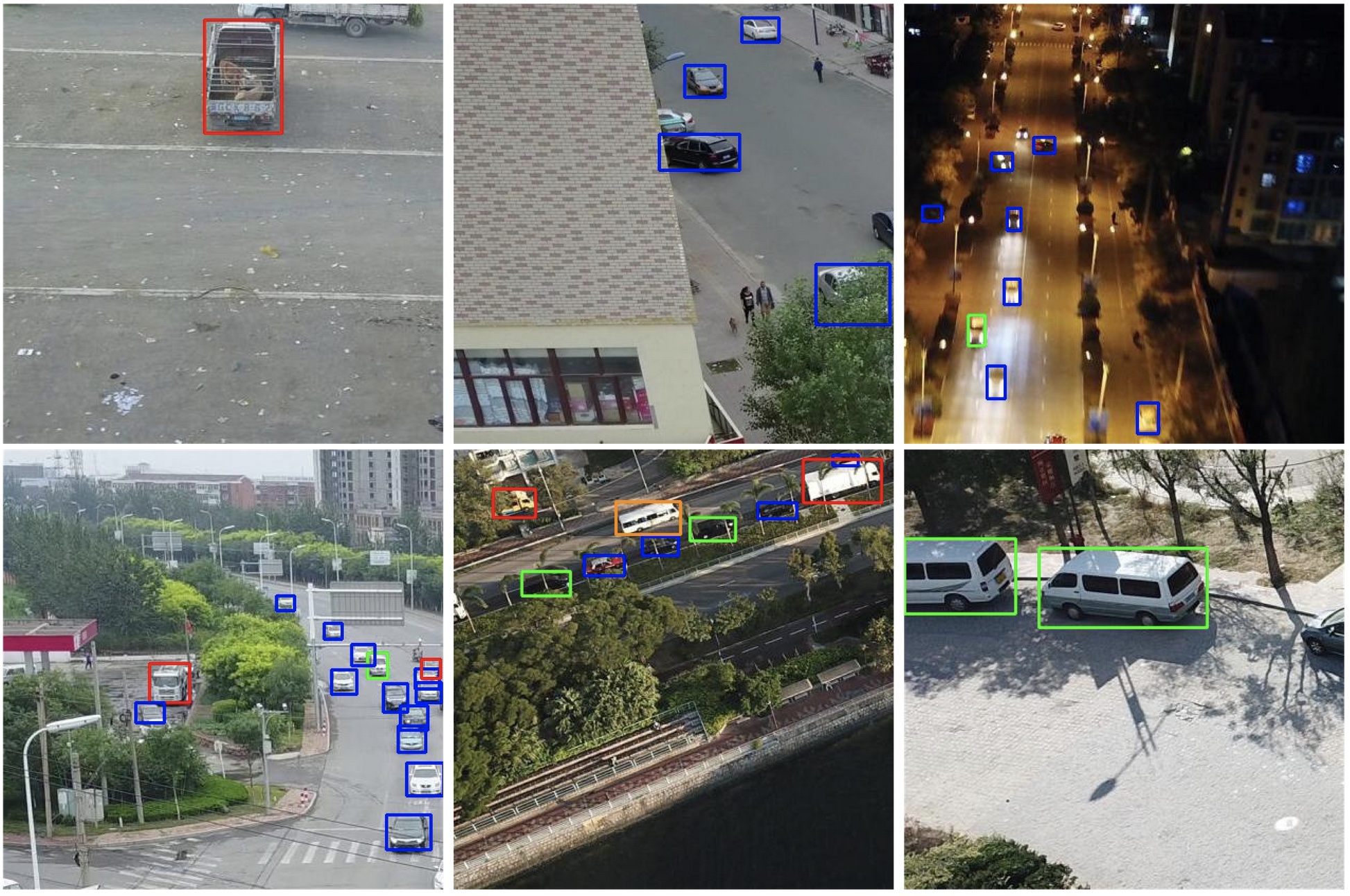}
\end{center}
  %\vspace{-5pt}
   \caption{Training image chips for our 4-class model. Ground truth bounding box colors correspond to Figure \ref{fig:histo_filt}.}
\label{fig:grid0}
\end{figure}

In this paper we focus on obfuscating vehicles, so we retain only labels for four object classes (see Figure \ref{fig:histo_filt}): bus, car, truck, and van.  This filtering of labels simplifies both the object detection and camouflage tasks.  We tile the VisDrone training imagery to $416 \times 416$ pixel windows for ease of ingestion into the YOLO \cite{yolov3} object detection family, see Figure \ref{fig:grid0}.

\section{Adversarial Camouflage}\label{sec:algo}

\subsection{Object Detector}

\begin{figure}%[t]
\begin{center}
   \includegraphics[width=0.9\linewidth]{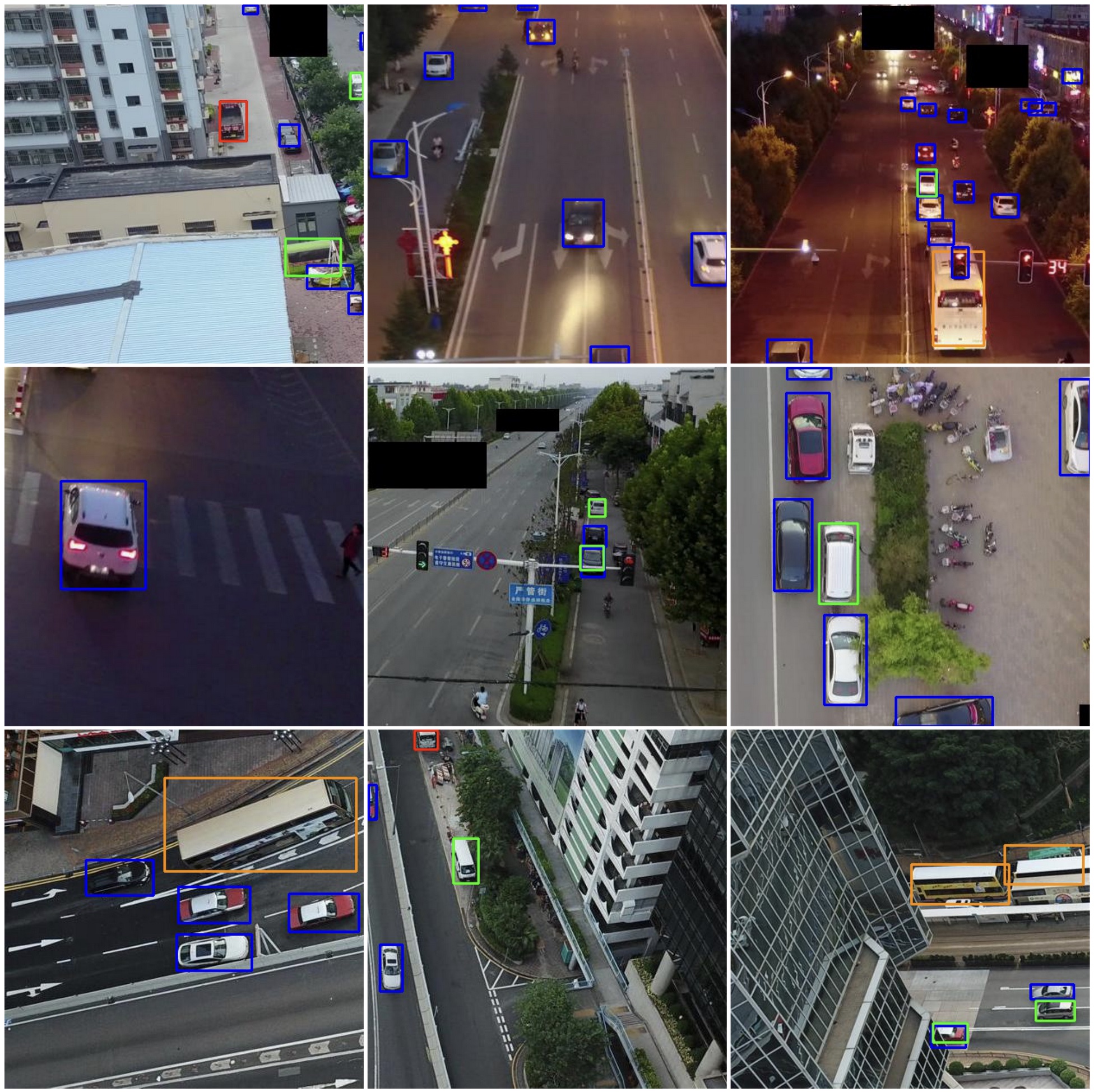}
\end{center}
  %\vspace{-5pt}
   \caption{Predictions for the 4-class YOLT model (bus = yellow, car = blue, truck = red, van = green).}
\label{fig:yolt_preds}
\end{figure}

We use the YOLTv4 \cite{yoltv4} object detection framework to train a 4-class vehicle detector.  We use a configuration file with an output feature map of $26\times26$ for improved detection of small objects.  
 % 60,000 batches, batch=64, 44493 images in train set, so 86 epochs.
 % local_data/cosmiq/src/avanetten/yoltv4/darknet$ cat cfg/_backup/yolt2_ave_26x26_visdrone_v1_4cat.cfg 
 % cat /local_data/cosmiq/wdata/avanetten/VisDrone/data/VisDrone2019-DET-train/yolt/v1_4cat/txt/visdrone_v1_4cat_train.data
 We train for 80 epochs using stochastic gradient descent, with a learning rate of 0.001, and a momentum of 0.9. Predictions with the trained model are shown in Figure \ref{fig:yolt_preds}.  We evaluate YOLT detection model performance with the 1610 images in the VisDrone test set, setting a true positive as a prediction of the correct class with an IOU$\geq0.5$.  Scores are shown in Table \ref{tab:yolt}; we report $1\sigma$ errors calculated via bootstrapping.
 
%# f1s_by_class: [0.5891989231445625, 0.7204086863479194, 0.43384729573432096, 0.4576274549113658]
%# mF1_score: 0.5502705900345422

% 	idx 	mAP 	mF1 	bus_AP 	bus_F1 	car_AP 	car_F1 	truck_AP 	truck_F1 	van_AP 	van_F1
%50 	mean 0.464645 	0.551205 	0.497359 	0.588085 	0.694220 	0.720768 	0.337506 	0.436150 	0.329496 	0.459817
%51 	std 	0.005072 	0.003687 	0.009741 	0.006824 	0.004202 	0.002304 	0.012640 	0.009114 	0.006478 	0.005452

\begin{table}[h]
  \caption{YOLT 4-class Detection Performance}
  \vspace{2pt}
  \label{tab:yolt}
  \small
  \centering
   \begin{tabular}{ll}
    \hline
    %\toprule
    % {\bf Dataset} & {\bf GSD (m)} & {\bf Object Size (m / pix)}  & {\bf F1} \\
    {\bf Class} & {\bf F1} \\
    \hline
    {Bus} & $0.59\pm0.06$ \\
    {Car} & $0.72\pm0.002$ \\
    {Truck} & $0.43\pm0.009$ \\
    {Van} &  $0.46\pm0.005$ \\
    \hline
    {Mean} & $0.55\pm0.003$ \\
   \end{tabular}
  \label{tab:comp}
  %\vspace{-5pt}
\end{table}

%%%%%%%%%%
\begin{comment}
\begin{table}[h]
  \caption{YOLTv4 4-class Detection Performance}
  \vspace{1pt}
  \label{tab:yolt}
  \small
  \centering
   \begin{tabular}{llllll}
    \hline
    %\toprule
    % {\bf Dataset} & {\bf GSD (m)} & {\bf Object Size (m / pix)}  & {\bf F1} \\
   &  {\bf Bus} & {\bf Car} & {\bf Truck}  & {\bf Van} & {\bf (Mean)}\\
    \hline
    {\bf F1} & $0.59\pm0.06$ & $0.72\pm0.002$ & $0.43\pm0.009$ & $0.46\pm0.005$ & $0.55\pm0.003$ \\
   \end{tabular}
  \label{tab:comp}
  %\vspace{-5pt}
\end{table}
\end{comment}
%%%%%%%%%%

\subsection{Adversarial Patches}\label{sec:patch}

To train an adversarial patch, we develop the Camolo \cite{camolo} codebase, which is a modification of the adversarial-yolo\footnote{\url{https://gitlab.com/EAVISE/adversarial-yolo}} codebase used in \cite{adv-yolo}.  The  adversarial-yolo codebase takes a trained model and labeled imagery as input, and attempts to create a patch that when overlaid on objects of interest will fool the detector.  Camolo makes a number of modifications:

\begin{enumerate}
	\item{Increased flexibility with input variables (\eg target patch size)}
	\item{Use with more recent versions of YOLO}
	\item{Allow patches to be semi-translucent}
\end{enumerate}	

\begin{figure}%[t]
\begin{center}
   \includegraphics[width=0.9\linewidth]{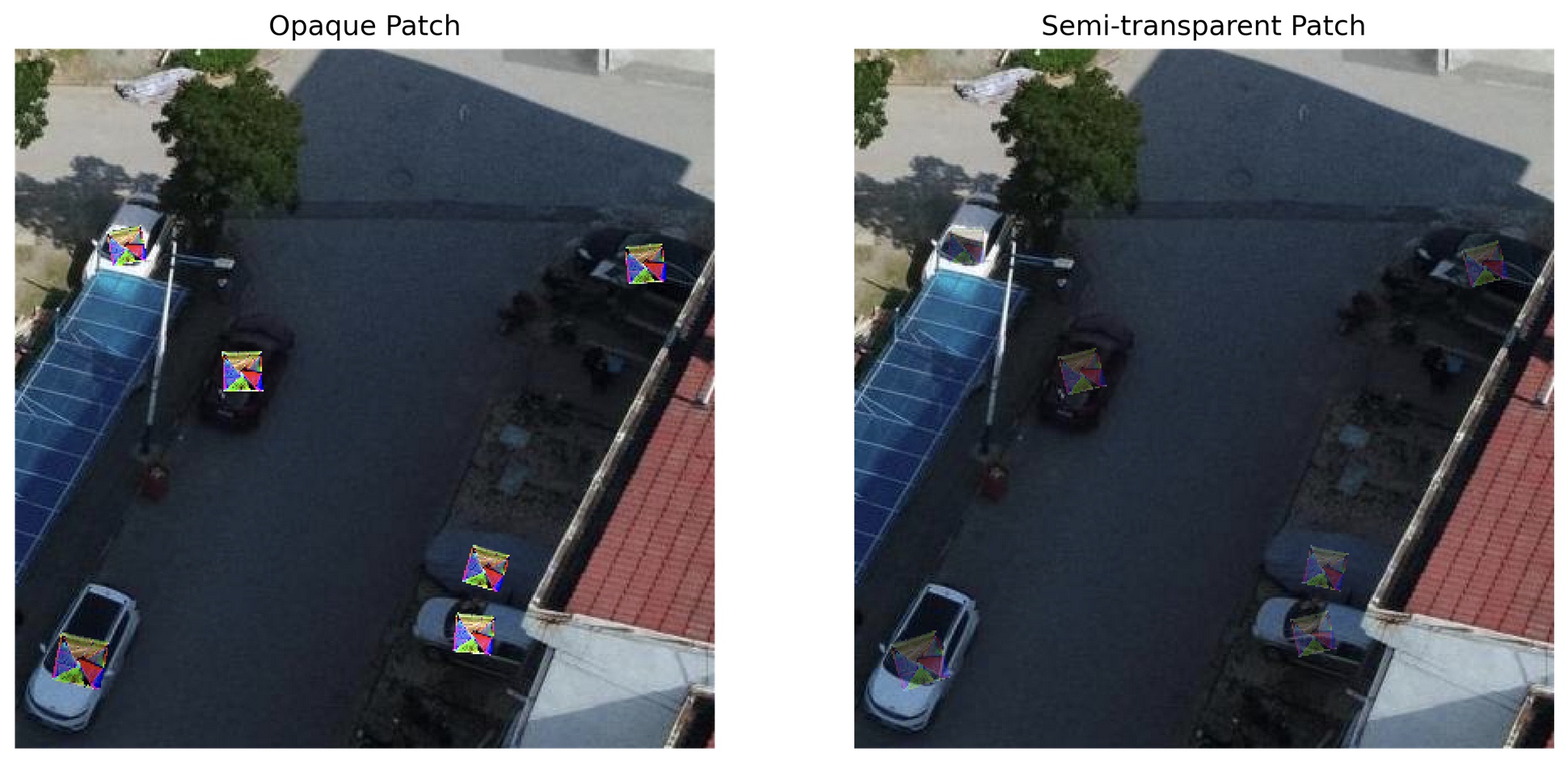}
\end{center}
  %\vspace{-5pt}
   \caption{Sample adversarial patch from \cite{adv-yolo} overlaid (left = opaque, right = semi-translucent) on VisDrone imagery. 
   % {\bf Left}: standard opaque overlay. {\bf Right}: semi-translucent overlay.
   }
\label{fig:alpha_patches}
\end{figure}

The most significant change (\#3) is the method of overlaying patches according to a selected alpha value, which dictates how transparent the patch appears. We postulate that a semi-translucent patch may help camouflage the patches themselves.  Previous studies have simply overwritten the existing pixels in an image with the desired patch.  We combine the patch and original image pixels according to a desired alpha value of the patch (alpha = 1 corresponds to an opaque patch, with alpha = 0 yielding an invisible patch). In Figure \ref{fig:alpha_patches} we overlay a sample adversarial patch on VisDrone imagery with both the standard fully opaque method, as well as semi-translucent.
 
\section{Experiments}\label{sec:experiments}

\subsection{Adversarial Patch Generation}\label{sec:patches}

\begin{figure}%[t]
\begin{center}
   \includegraphics[width=0.98\linewidth]{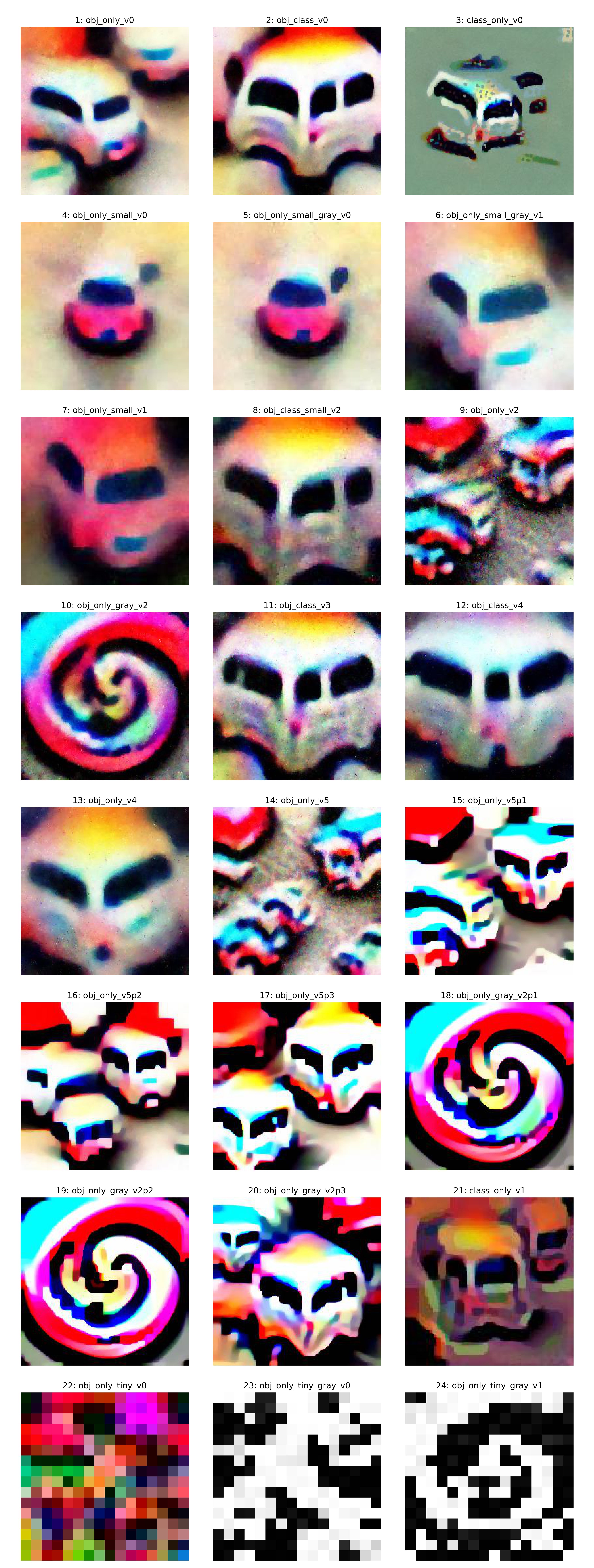}
\end{center}
  %\vspace{-5pt}
   \caption{Trained adversarial patch library. 
   % {\bf Left}: standard opaque overlay. {\bf Right}: semi-translucent overlay.
   }
\label{fig:patch_grid}
\end{figure}

We train a variety of adversarial patches using the Camolo codebase and VisDrone dataset.  All experiments use the same initial dataset and model architecture.  We vary the starting patch between experiments, trying both legacy patches as well as totally random starting points.  Other variables are the allowed colors of the patches, and the alpha value (translucency) of the patches.  The patch size (as a fraction of the area of the object of interest), and noise level is also varied.  Finally, we select one of three losses for each experiment: object (focus only on minimizing bounding box detections), class (focus on confusing the class prediction of each bounding box), and object $\times$ class. Training occurs for a minimum of 40 epochs for each experiment.  Parameters are shown in Table \ref{tab:exps}.  
Most experiments aim to yield a non-detection (\eg 1. obj\_only\_v0)) though
some experiments seek to confuse which object is classified (\eg 3. class\_only\_v0), such as classifying a car as a truck.  See Figure \ref{fig:patch_grid} for the trained patches.
%Most experiments seek only to obfuscate the bounding box of our category of interest: car.  
%Some experiments, also seek to confuse which object is classified (\eg 2. class\_only\_v0), such as classifying a car as a truck.  See Figure \ref{fig:patch_grid} for the trained patches.

%mF1 (camo applied) mean: 0.3842731433672027 std 0.08394210011967035
%F1 (car) mean: 0.6322758000861727 std 0.05148778246369784
%F1 (patch detection) mean: 0.6846381555942238 std 0.2139977860105053
%mF1 % Reduction mean: 33.14537132655946 std 16.78842002393407

\begin{table}[h]
  \caption{Adversarial Patches Trained on VisDrone}
  \vspace{2pt}
  \label{tab:patches}
  \small
  \centering
   \begin{tabular}{lllll}
    \hline
 & {\bf Name} & {\bf Loss} & {\bf Size} & {\bf Alpha} \\
 \hline
1 & obj\_only\_v0 & obj & $0.2$ & $0.4$ \\
2 & obj\_class\_v0 & obj * cls & $0.2$ & $0.4$ \\
3 & class\_only\_v0 & cls & $0.2$ & $0.4$ \\
4 & obj\_only\_small\_v0 & obj & $0.2$ & $0.3$ \\
5 & obj\_only\_small\_gray\_v0 & obj & $0.2$ & $0.3$ \\
6 & obj\_only\_small\_gray\_v1 & obj & $0.16$ & $0.4$ \\
7 & obj\_only\_small\_v1 & obj & $0.16$ & $0.6$ \\
8 & obj\_class\_small\_v2 & obj * cls & $0.16$ & $0.75$ \\
9 & obj\_only\_v2 & obj & $0.25$ & $0.8$ \\
10 & obj\_only\_gray\_v2 & obj & $0.25$ & $0.8$ \\
11 & obj\_class\_v3 & obj * cls & $0.16$ & $0.6$ \\
12 & obj\_class\_v4 & obj * cls & $0.12$ & $0.5$ \\
13 & obj\_only\_v4 & obj & $0.12$ & $0.5$ \\
14 & obj\_only\_v5 & obj & $0.3$ & $0.5$ \\
15 & obj\_only\_v5p1 & obj & $0.25$ & $0.5$ \\
16 & obj\_only\_v5p2 & obj & $0.25$ & $0.4$ \\
17 & obj\_only\_v5p3 & obj & $0.2$ & $0.4$ \\
18 & obj\_only\_gray\_v2p1 & obj & $0.25$ & $0.7$ \\
19 & obj\_only\_gray\_v2p2 & obj & $0.25$ & $0.6$ \\
20 & obj\_only\_gray\_v2p3 & obj & $0.2$ & $0.7$ \\
21 & class\_only\_v1 & cls & $0.25$ & $0.9$ \\
22 & obj\_only\_tiny\_v0 & obj & $0.25$ & $1.0$ \\
23 & obj\_only\_tiny\_gray\_v0 & obj & $0.25$ & $1.0$ \\
24 & obj\_only\_tiny\_gray\_v1 & obj & $0.2$ & $0.75$ \\
   \end{tabular}
  \label{tab:exps}
  %\vspace{-5pt}
\end{table}

\begin{figure}
\begin{tabular}{c}
  \includegraphics[width=0.95\linewidth] {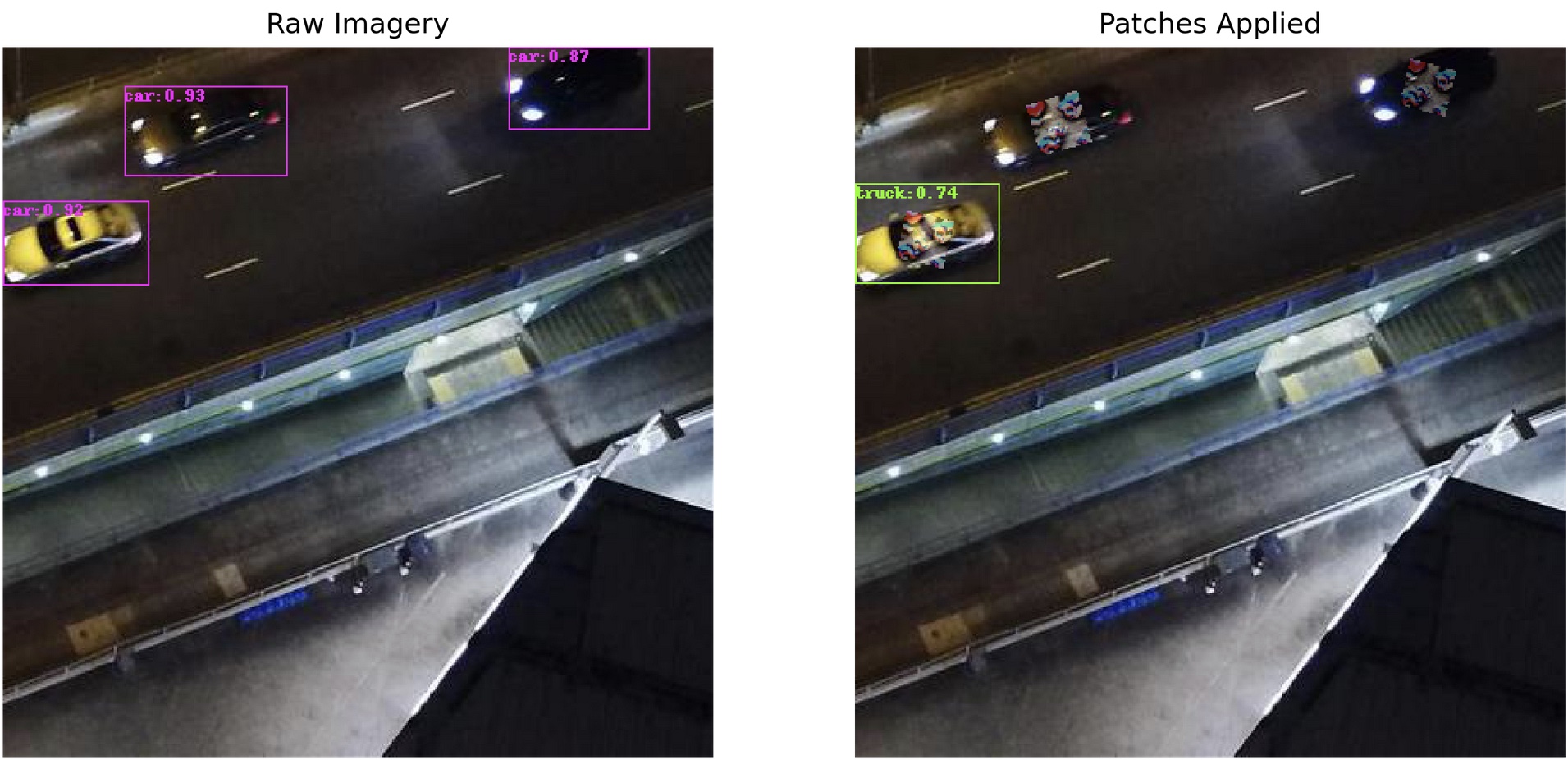} \\  
  [-2pt] 
(a) 14. obj\_only\_v5 \\  [4pt] 
 \includegraphics[width=0.95\linewidth]{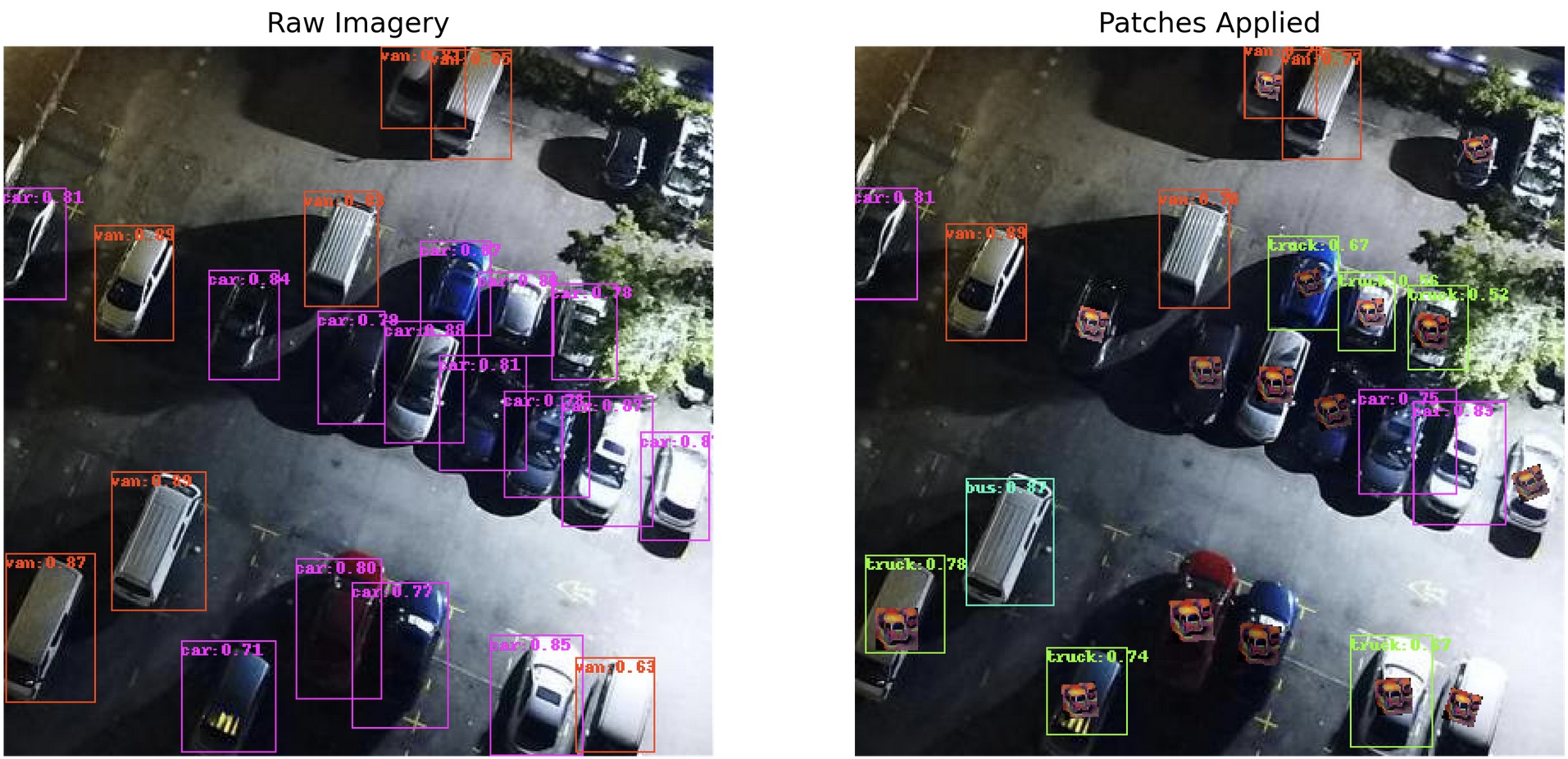} \\  
(b) 21. class\_only\_v1 \\  [-0pt] 
% \multicolumn{2}{c}{\includegraphics[width=0.99\linewidth]{it} }\\
% \multicolumn{2}{c}{(e) fifth}
 \includegraphics[width=0.95\linewidth]{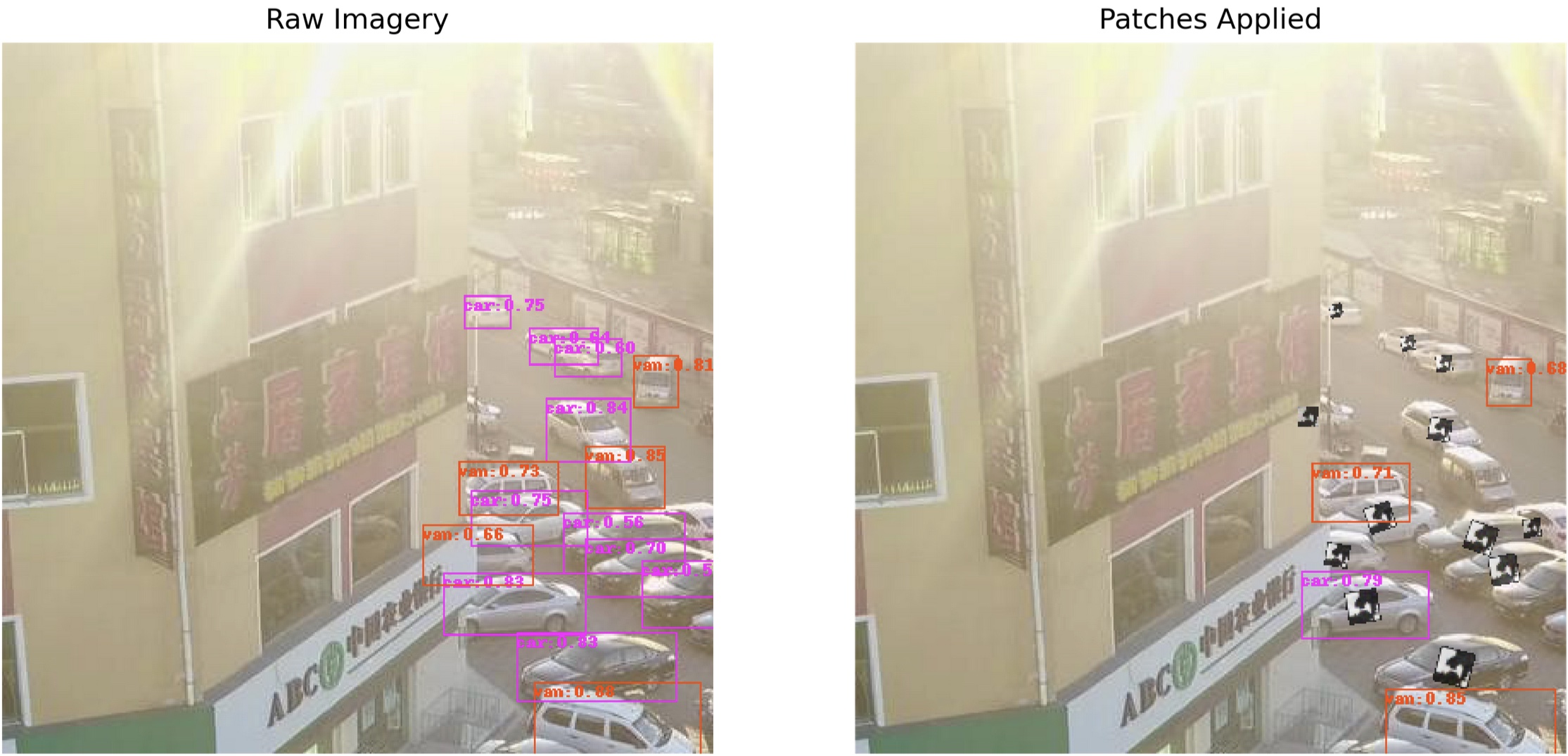} \\  
(c) 24. obj\_only\_tiny\_gray\_v1 \\  [-0pt] 

\end{tabular}
\vspace{2pt}
\caption{Detection of vehicles with the original trained model with raw test imagery (left) and patches applied on cars (right).}
\label{fig:foolin}
%\vspace{-1pt}
\end{figure}

\begin{figure}
\begin{tabular}{ll}
  \includegraphics[width=0.5\linewidth] {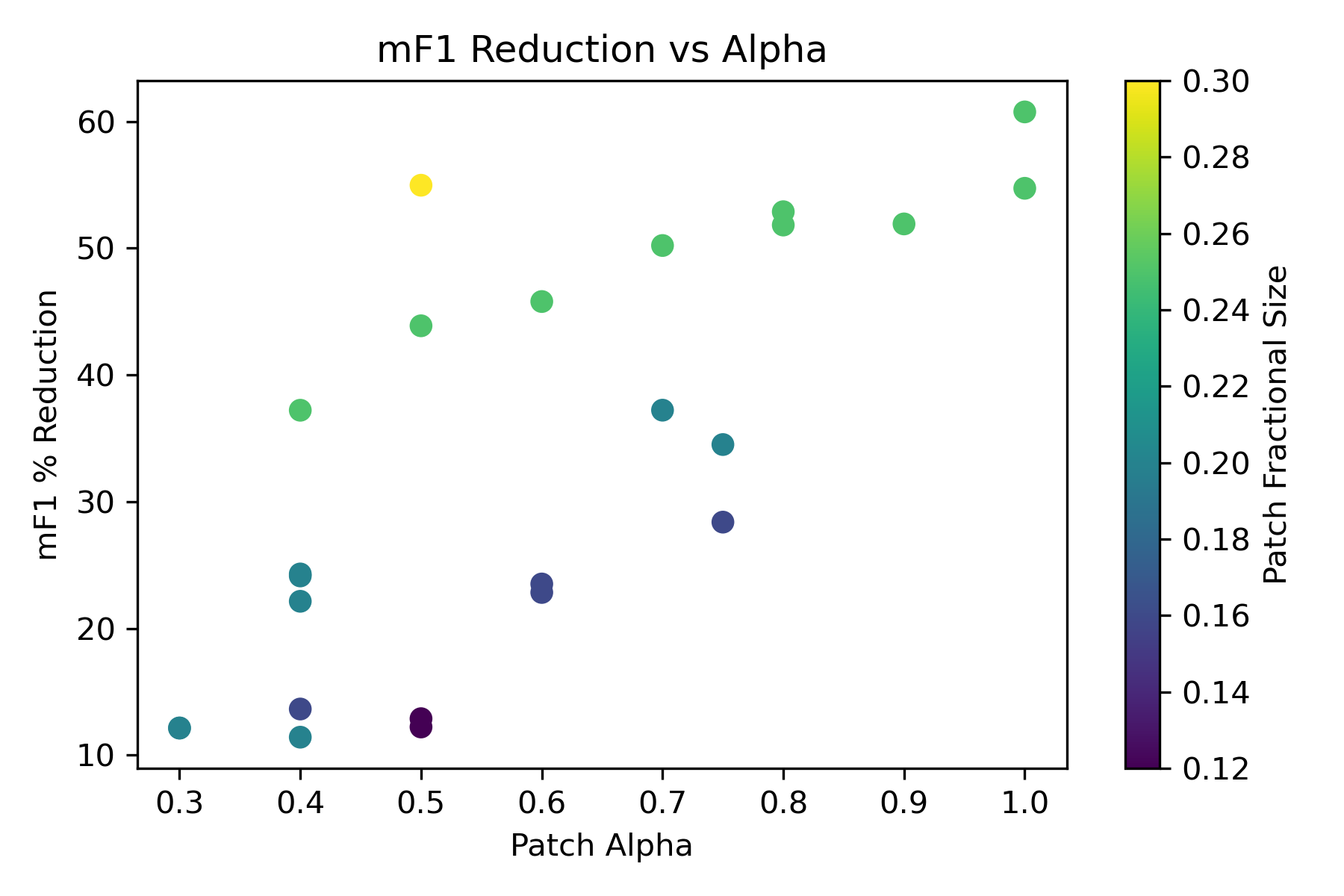} & \includegraphics[width=0.5\linewidth]{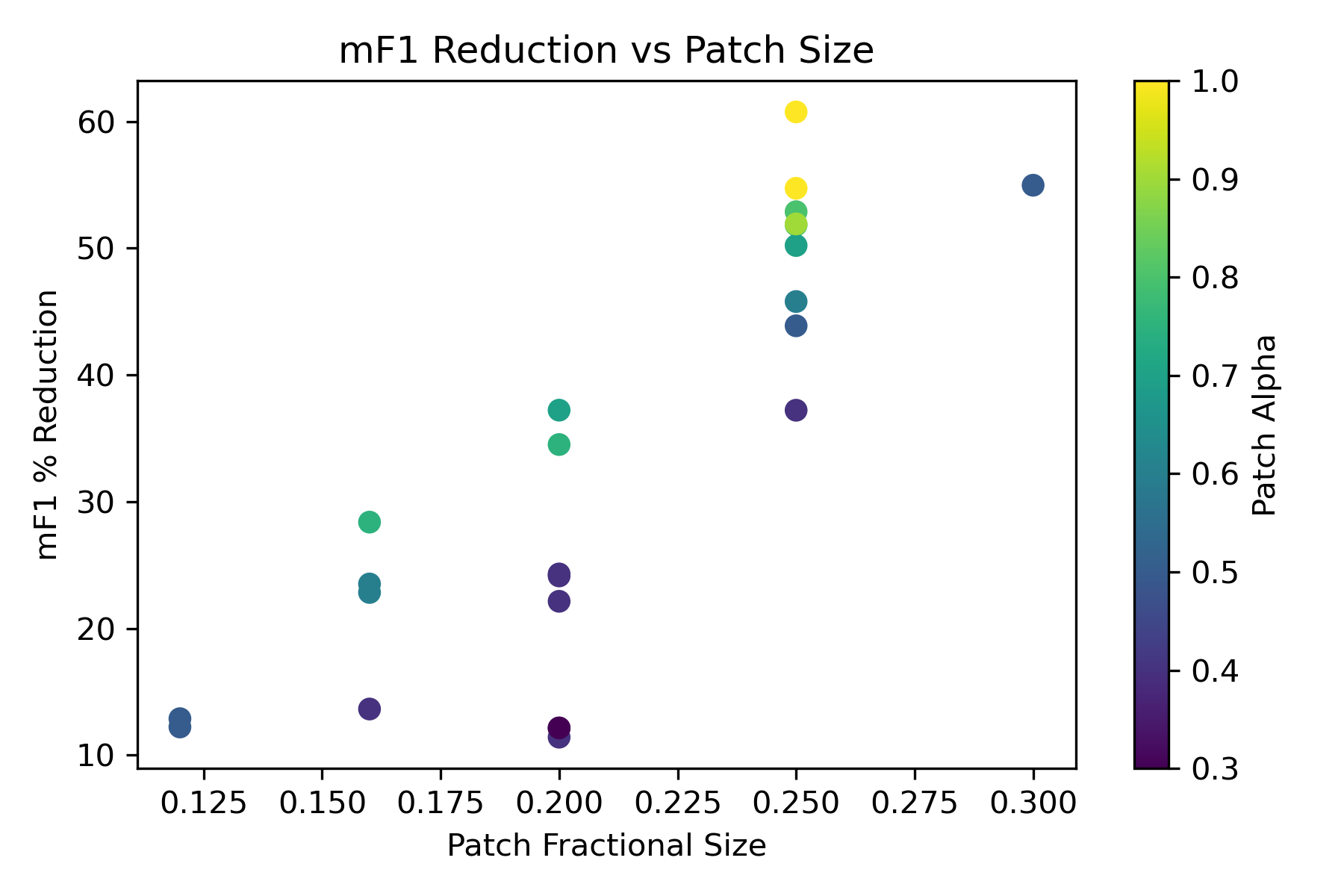} \\  
  [-2pt] 
\end{tabular}
\caption{Patch performance as a function of size and alpha (higher is better). }
\label{fig:f1_plots}
%\vspace{-1pt}
\end{figure}

In Figure \ref{fig:foolin} we show successful examples of  patches fooling our trained object detector. 
Figure \ref{fig:f1_plots} shows how mF1 varies with alpha (translucency) and patch size.
This plot shows the percentage reduction in vehicle detection provided by the patches; the Pearson correlation coefficient between vehicle detection mF1 reduction and size is 0.83 and the correlation coefficient between alpha and mF1 reduction is 0.76.  

%\begin{comment}
%\end{comment}

\subsection{Detection of Patches}\label{sec:patch_detections}

\begin{figure}%[t]
\begin{center}
   \includegraphics[width=0.999\linewidth]{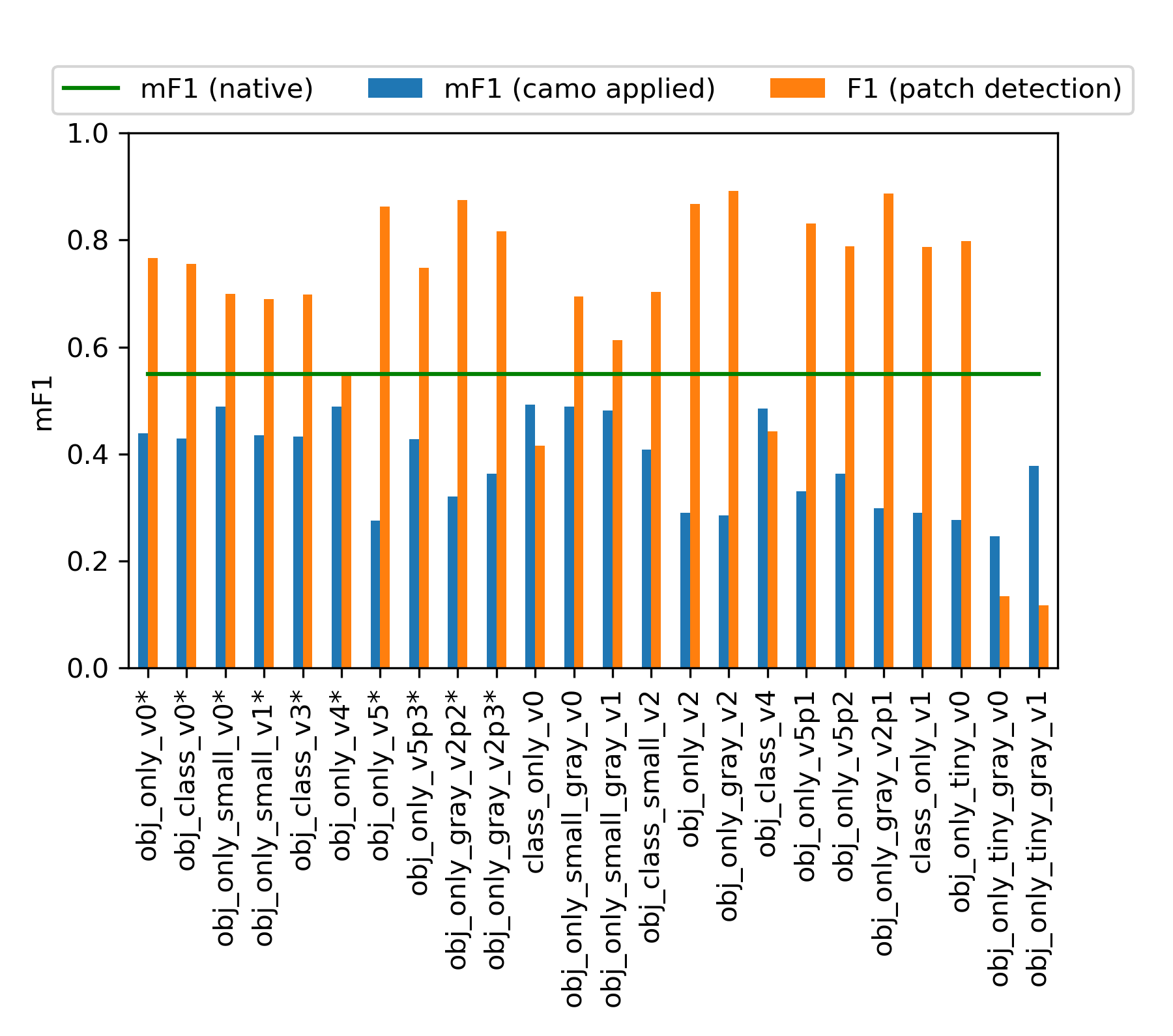}
\end{center}
  \vspace{-5pt}
   \caption{Detection performance for each model. 
   }
\label{fig:f1_bar}
\end{figure}

We showed in Figure \ref{fig:f1_plots}  that our patches significantly reduce the ability of our trained YOLTv4 model.  In this section, we explore how easy it is to detect the existence of these patches. Recall in the introduction that we found legacy patches to be detectable with nearly perfect performance.  We train a generic patch detector using the same raw VisDrone imagery as in Section \ref{sec:patches}.  We select 10 patches to train with, and overlay these patches on the training imagery.  We train a YOLTv4 model, and test our generic patch detector by overlaying each of our 24 patches on the test set and scoring how robustly the patches can be detected.  Performance is listed in Table \ref{tab:exps2}, and illustrated in Figure \ref{fig:f1_bar}.  In Figure  \ref{fig:f1_bar}, the green line denotes the detection performance of the original 4-class model; the blue bars denote the performance of the 4-class detection model when the listed adversarial patch has been applied; the orange bars denote the detection performance of the model trained to detect the existence of patches.  The 10 patches used to train the patch detection model are appended with an asterisk (\ie $*$) in Table \ref{tab:exps2} and  \ref{fig:f1_bar}. %; note that the orange bars are not significantly higher for asterisked patches compared to the ``unseen'' patches.  Also note that for most patches, it is easier to detect the existence of the patch than vehicles.  

\begin{table}[h]
  \caption{Detection Performance of Various Models}
  \vspace{4pt}
  \label{tab:patches}
  \small
  \centering
   \begin{tabular}{llll}
    \hline
 & {\bf Model Name} & {\bf mF1$_{camo}$} & {\bf F1$_{patch}$} \\
 \hline
1 & obj\_only\_v0* & $0.44$ & $0.77$ \\
2 & obj\_class\_v0* & $0.43$ & $0.76$ \\
3 & obj\_only\_small\_v0* & $0.49$ & $0.70$ \\
4 & obj\_only\_small\_v1* & $0.44$ & $0.69$ \\
5 & obj\_class\_v3* & $0.43$ & $0.70$ \\
6 & obj\_only\_v4* & $0.49$ & $0.55$ \\
7 & obj\_only\_v5* & $0.28$ & $0.86$ \\
8 & obj\_only\_v5p3* & $0.43$ & $0.75$ \\
9 & obj\_only\_gray\_v2p2* & $0.32$ & $0.87$ \\
10 & obj\_only\_gray\_v2p3* & $0.36$ & $0.82$ \\
11 & class\_only\_v0 & $0.49$ & $0.42$ \\
12 & obj\_only\_small\_gray\_v0 & $0.49$ & $0.69$ \\
13 & obj\_only\_small\_gray\_v1 & $0.48$ & $0.61$ \\
14 & obj\_class\_small\_v2 & $0.41$ & $0.70$ \\
15 & obj\_only\_v2 & $0.29$ & $0.87$ \\
16 & obj\_only\_gray\_v2 & $0.29$ & $0.89$ \\
17 & obj\_class\_v4 & $0.49$ & $0.44$ \\
18 & obj\_only\_v5p1 & $0.33$ & $0.83$ \\
19 & obj\_only\_v5p2 & $0.36$ & $0.79$ \\
20 & obj\_only\_gray\_v2p1 & $0.30$ & $0.89$ \\
21 & class\_only\_v1 & $0.29$ & $0.79$ \\
22 & obj\_only\_tiny\_v0 & $0.28$ & $0.80$ \\
23 & obj\_only\_tiny\_gray\_v0 & ${\bf 0.25} $ & $0.13$ \\
24 & obj\_only\_tiny\_gray\_v1 & $0.38$ & ${ \bf 0.12}$ \\
\hline
 & mean & 0.38 & 0.68 \\
 
%  car detection
% & {\bf Name} & {\bf F1$_{camo}$} & {\bf F1$_{patch}$} \\
%1 & obj\_only\_v0* & $0.66$ & $0.77$ \\
%2 & obj\_class\_v0* & $0.66$ & $0.76$ \\
%3 & obj\_only\_small\_v0* & $0.68$ & $0.7$ \\
%4 & obj\_only\_small\_v1* & $0.67$ & $0.69$ \\
%5 & obj\_class\_v3* & $0.67$ & $0.7$ \\
%6 & obj\_only\_v4* & $0.69$ & $0.55$ \\
%7 & obj\_only\_v5* & $0.55$ & $0.86$ \\
%8 & obj\_only\_v5p3* & $0.66$ & $0.75$ \\
%9 & obj\_only\_gray\_v2p2* & $0.6$ & $0.87$ \\
%10 & obj\_only\_gray\_v2p3* & $0.64$ & $0.82$ \\
%11 & class\_only\_v0 & $0.69$ & $0.42$ \\
%12 & obj\_only\_small\_gray\_v0 & $0.68$ & $0.69$ \\
%13 & obj\_only\_small\_gray\_v1 & $0.68$ & $0.61$ \\
%14 & obj\_class\_small\_v2 & $0.66$ & $0.7$ \\
%15 & obj\_only\_v2 & $0.59$ & $0.87$ \\
%16 & obj\_only\_gray\_v2 & $0.58$ & $0.89$ \\
%17 & obj\_class\_v4 & $0.69$ & $0.44$ \\
%18 & obj\_only\_v5p1 & $0.61$ & $0.83$ \\
%19 & obj\_only\_v5p2 & $0.62$ & $0.79$ \\
%20 & obj\_only\_gray\_v2p1 & $0.59$ & $0.89$ \\
%21 & class\_only\_v1 & $0.58$ & $0.79$ \\
%22 & obj\_only\_tiny\_v0 & $0.54$ & $0.8$ \\
%23 & obj\_only\_tiny\_gray\_v0 & $0.52$ & $0.13$ \\
%24 & obj\_only\_tiny\_gray\_v1 & $0.65$ & $0.12$ \\
  \end{tabular}
  \label{tab:exps2}
  %\vspace{-5pt}
\end{table}

\section{Analysis}\label{sec:analysis}

Note in Figure \ref{fig:f1_bar} that the orange bars are not significantly higher for asterisked (\ie training) patches compared to the ``unseen'' patches.  Also note that for most patches, it is easier to detect the existence of the patch than vehicles.  

\begin{figure}%[t]
\begin{center}
   \includegraphics[width=0.999\linewidth]{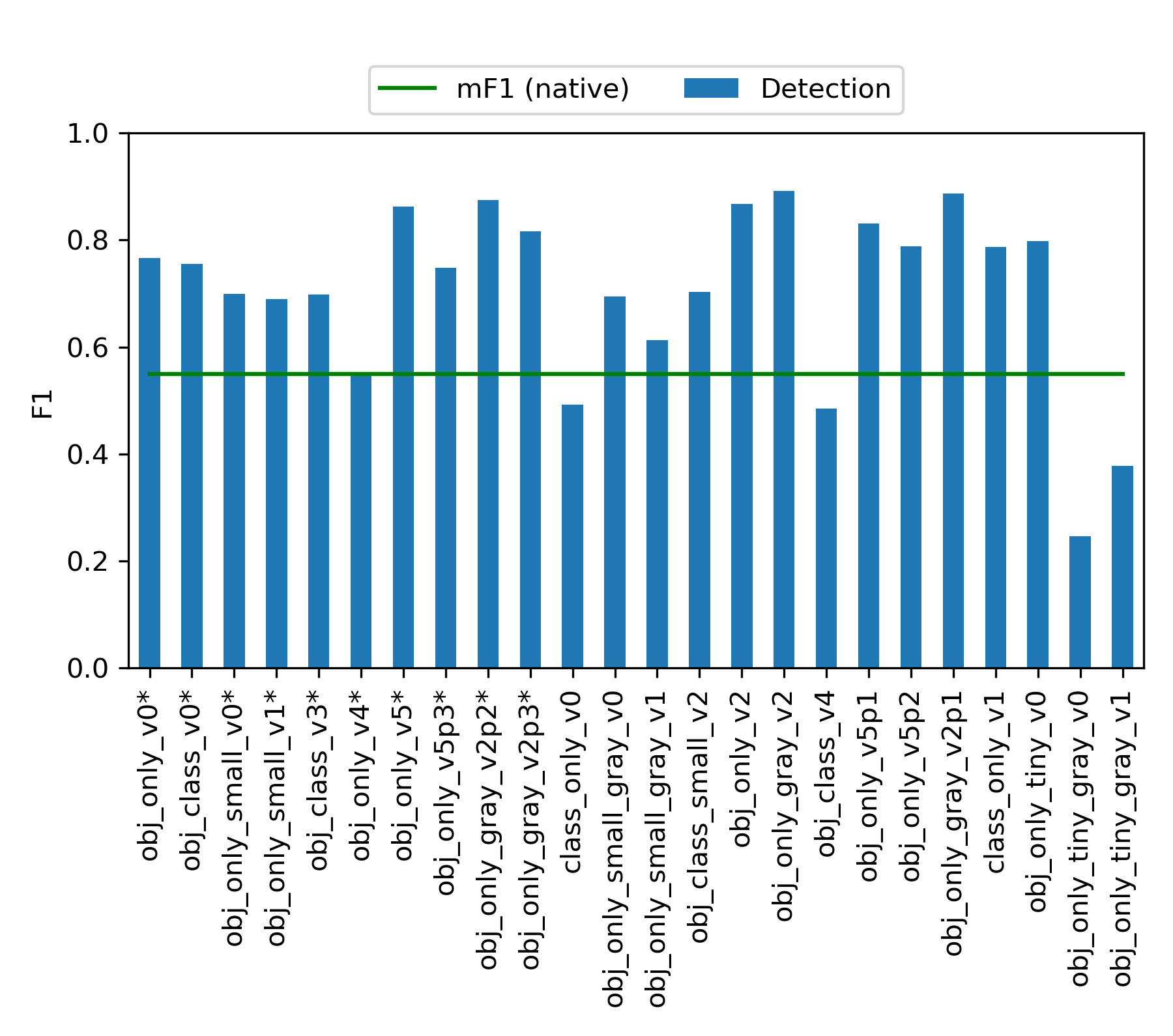}
\end{center}
  \vspace{-5pt}
   \caption{Aggregate detection performance for each model.  The lower the value, the more effective the patch. }
\label{fig:det}
\end{figure}

If we collapse the two performance of the two experiment groups (vehicle detection + patch detection), we are left with Figure \ref{fig:det}, which shows a ``detection`` score: maximum of the two blue and orange bars in  \ref{fig:f1_bar}.  This ``detection'' score provides a measure of the efficacy of the patch, since an easily detected patch is not terribly effective at %obfuscating the desired object.  
camouflage, since the patch itself exposes the presence of the object of interest.  
Recall that in Figures \ref{fig:f1_bar} and \ref{fig:det}, lower is better.  Note also that most patches provide no aggregate benefit.  Yet the two black and white patches (obj\_only\_tiny\_gray\_v0 and obj\_only\_tiny\_gray\_v1) are the most effective.  The precise reason for this efficacy will be left to later work, but we postulate that the marked difference of these two patches (\ie grayscale vs color) from the ten patches in our patch detector training set is largely responsible.  

\begin{figure}
\vspace{-1pt}
\begin{tabular}{ll}
  \includegraphics[width=0.5\linewidth] {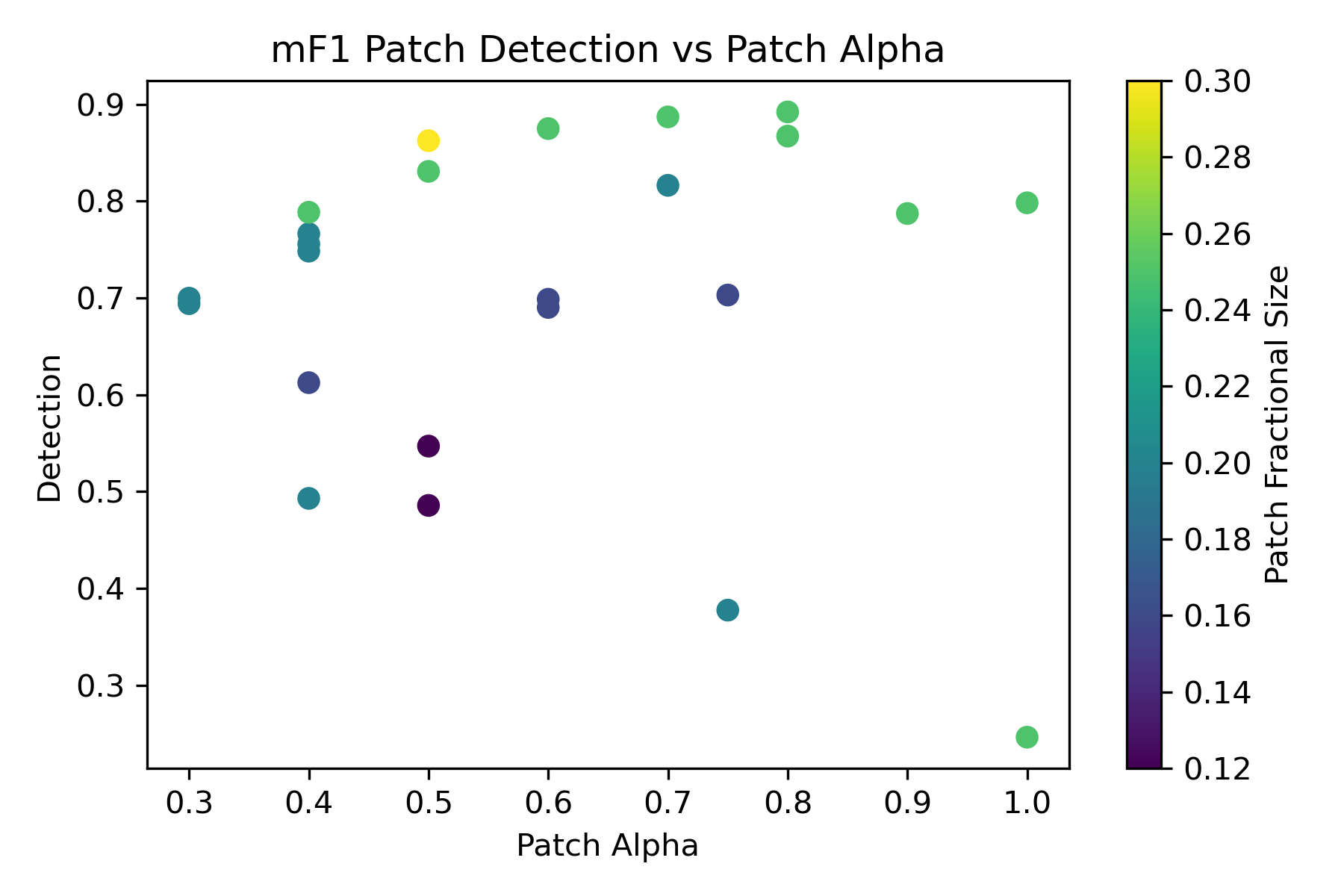} & \includegraphics[width=0.5\linewidth]{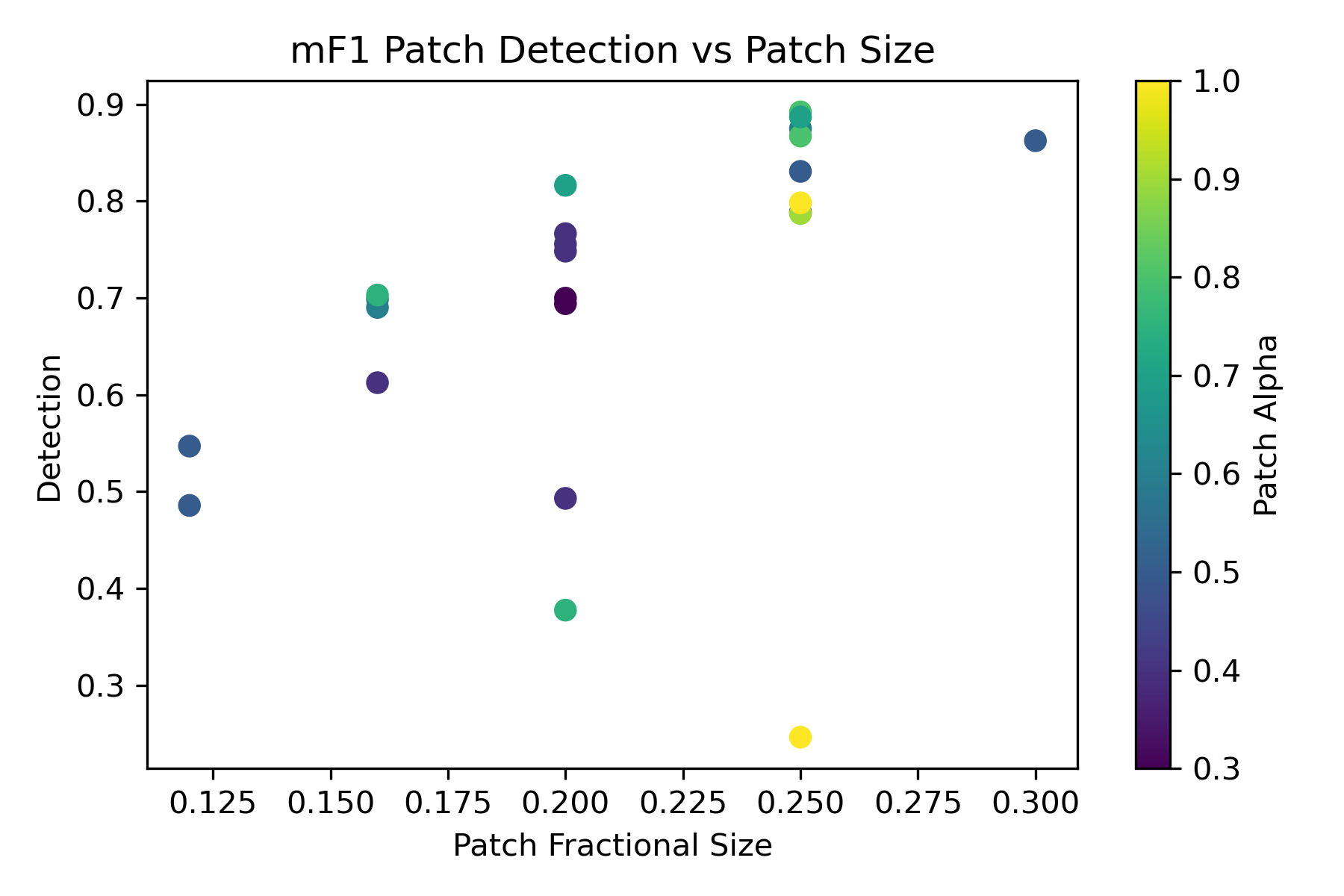} \\  

%   \includegraphics[width=0.48\linewidth] {figs/f1_vs_alpha_car.png} & \includegraphics[width=0.48\linewidth]{figs/f1_vs_size_car.png} \\  
  %\vspace{-6pt} 
\end{tabular}
  \vspace{-3pt} 
\caption{Detection dependence on size, alpha (lower is better).
%Top: mF1, Bottom: total detection (max of vehicle detection and patch detection). 
}
\label{fig:f1_plots2}
%\vspace{-1pt}
\end{figure}

Figure \ref{fig:f1_plots2} shows how aggregate varies with alpha and patch size. Recall from Figure \ref{fig:f1_plots} that larger and more opaque (higher alpha) patches are more effective at confusing the vehicle detector.  Yet  Figure \ref{fig:f1_plots2} shows that patches with higher alpha and larger sizes are actually less effective in aggregate performance since the existence of large, opaque patches is far easier to detect.  In fact, in terms of aggregate performance, smaller, more translucent patches are preferred (Pearson correlation between detection and alpha: -0.76, Pearson correlation between detection and patch size: -0.83).   Ultimately, for true camouflage, one should apparently prioritize patch stealthiness.
% This plot demonstrates that the performance of the patches is highly correlated with both the patch size and alpha (translucency).  The top row of  \ref{fig:f1_plots} shows percentage reduction in vehicle detection provided by the patches; the Pearson correlation coefficient between vehicle detection mF1 reduction and size is 0.83 and the correlation coefficient between alpha and mF1 reduction is 0.76.  The bottom row illustrates the aggregate detection performance of the patches

\section{Conclusions}

Adversarial patches have been shown to be effective in camouflaging objects in relatively homogeneous datasets such as Inria and DOTA.  In this paper we showed that while legacy patches may be effective in hiding people and aircraft, such patches are trivial to detect.  This motivates our study of whether ``stealthy'' patches can be designed to obfuscate objects in overhead imagery.  Using the diverse VisDrone dataset we train a library of 24 adversarial patches with various input parameters.  While most of these patches significantly reduce the detection of our objects of interest (buses, cars, trucks, vans), most patches are still easier to detect than vehicles. Our two black-and-while patches are poorly detected with our patch detection model, however, due to their significant variance from the patch training set.  This raises the question: how large and diverse of a patch library is required to be truly effective?  And how much effort is required on the mitigation side in order to train a robust patch detection model that will effectively combat adversarial camouflage?  We have provided some first hints to these questions, but much work remains to be done.  Besides diving into these questions, in future work we hope to introduce false positives into imagery (\eg can a simple pattern laid out in an empty field trick a computer vision model into ``detecting'' a full parking lot).

{\small
\bibliographystyle{ieee_fullname}
\bibliography{camo_final}

\begin{thebibliography}{10}\itemsep=-1pt

\bibitem{aerial_camo}
Ajaya Adhikari, Richard J.~M. den Hollander, Ioannis Tolios, Michael van
  Bekkum, Anneloes Bal, Stijn Hendriks, Maarten Kruithof, Dennis Gross, Nils
  Jansen, Guillermo~A. P{\'{e}}rez, Kit Buurman, and Stephan Raaijmakers.
\newblock Adversarial patch camouflage against aerial detection.
\newblock {\em CoRR}, abs/2008.13671, 2020.

\bibitem{inria}
N. Dalal and B. Triggs.
\newblock Histograms of oriented gradients for human detection.
\newblock In {\em 2005 IEEE Computer Society Conference on Computer Vision and
  Pattern Recognition (CVPR'05)}, volume~1, pages 886--893 vol. 1, 2005.

\bibitem{nat_style_camo}
Ranjie Duan, Xingjun Ma, Yisen Wang, James Bailey, A.~Kai Qin, and Yun Yang.
\newblock Adversarial camouflage: Hiding physical-world attacks with natural
  styles.
\newblock {\em CoRR}, abs/2003.08757, 2020.

\bibitem{adv_laser}
Ranjie Duan, Xiaofeng Mao, A.~Kai Qin, Yun Yang, Yuefeng Chen, Shaokai Ye, and
  Yuan He.
\newblock Adversarial laser beam: Effective physical-world attack to dnns in a
  blink.
\newblock {\em CoRR}, abs/2103.06504, 2021.

\bibitem{yoltv4}
Adam~Van Etten.
\newblock Announcing yoltv4: Improved satellite imagery object detection.
\newblock \url{https://medium.com/@avanetten}, 2021.

\bibitem{camolo}
Adam~Van Etten.
\newblock Camolo.
\newblock \url{https://github.com/IQTLabs/camolo}, 2022.

\bibitem{robust_attacks}
Ivan Evtimov, Kevin Eykholt, Earlence Fernandes, Tadayoshi Kohno, Bo Li, Atul
  Prakash, Amir Rahmati, and Dawn Song.
\newblock Robust physical-world attacks on machine learning models.
\newblock {\em CoRR}, abs/1707.08945, 2017.

\bibitem{univ_pert}
Jan {Hendrik Metzen}, Mummadi {Chaithanya Kumar}, Thomas {Brox}, and Volker
  {Fischer}.
\newblock {Universal Adversarial Perturbations Against Semantic Image
  Segmentation}.
\newblock {\em arXiv e-prints}, page arXiv:1704.05712, Apr. 2017.

\bibitem{adv_veh}
Jiajun Lu, Hussein Sibai, Evan Fabry, and David~A. Forsyth.
\newblock {NO} need to worry about adversarial examples in object detection in
  autonomous vehicles.
\newblock {\em CoRR}, abs/1707.03501, 2017.

\bibitem{yolo}
Joseph Redmon, Santosh~Kumar Divvala, Ross~B. Girshick, and Ali Farhadi.
\newblock You only look once: Unified, real-time object detection.
\newblock {\em CoRR}, abs/1506.02640, 2015.

\bibitem{yolov3}
Joseph Redmon and Ali Farhadi.
\newblock Yolov3: An incremental improvement.
\newblock {\em arXiv}, 2018.

\bibitem{rareplanes}
Jacob Shermeyer, Thomas Hossler, Adam~Van Etten, Daniel Hogan, Ryan Lewis, and
  Daeil Kim.
\newblock Rareplanes: Synthetic data takes flight.
\newblock {\em CoRR}, abs/2006.02963, 2020.

\bibitem{adv-yolo}
Simen Thys, Wiebe~Van Ranst, and Toon Goedem{\'{e}}.
\newblock Fooling automated surveillance cameras: adversarial patches to attack
  person detection.
\newblock {\em CoRR}, abs/1904.08653, 2019.

\bibitem{attn_camo}
Jiakai Wang, Aishan Liu, Zixin Yin, Shunchang Liu, Shiyu Tang, and Xianglong
  Liu.
\newblock Dual attention suppression attack: Generate adversarial camouflage in
  physical world.
\newblock {\em CoRR}, abs/2103.01050, 2021.

\bibitem{dota}
Gui{-}Song Xia, Xiang Bai, Jian Ding, Zhen Zhu, Serge~J. Belongie, Jiebo Luo,
  Mihai Datcu, Marcello Pelillo, and Liangpei Zhang.
\newblock {DOTA:} {A} large-scale dataset for object detection in aerial
  images.
\newblock {\em CoRR}, abs/1711.10398, 2017.

\bibitem{adv_2017}
Cihang Xie, Jianyu Wang, Zhishuai Zhang, Yuyin Zhou, Lingxi Xie, and Alan~L.
  Yuille.
\newblock Adversarial examples for semantic segmentation and object detection.
\newblock {\em CoRR}, abs/1703.08603, 2017.

\bibitem{visdrone}
Pengfei Zhu, Longyin Wen, Dawei Du, Xiao Bian, Heng Fan, Qinghua Hu, and Haibin
  Ling.
\newblock Detection and tracking meet drones challenge.
\newblock {\em IEEE Transactions on Pattern Analysis and Machine Intelligence},
  pages 1--1, 2021.

\end{thebibliography}
}

\end{document}